\documentclass[12pt, a4paper]{article}
\usepackage{geometry}
\usepackage{graphicx} 

\graphicspath{{Figures/}{./}} 
\usepackage{subcaption}
\usepackage{array}
\usepackage{amssymb}
\usepackage{amsmath}
\usepackage{amsfonts}
\numberwithin{equation}{section}
\usepackage{bbm}
\usepackage{bm}
\usepackage{booktabs}
\usepackage{authblk}
\usepackage{latexsym}
\usepackage{enumitem}
\usepackage{hyperref}

\hypersetup{
    colorlinks=true,
    linkcolor=blue,
    filecolor=magenta,      
    urlcolor=cyan
    }

\usepackage[sorting=none,sortcites=true,style=ieee,url=false,giveninits=true]{biblatex}
\AtEveryBibitem{\clearlist{language}}
\AtEveryBibitem{%
  \clearfield{note}%
}
\title{A metrological framework for uncertainty evaluation in machine learning classification models}
\author[1]{Samuel Bilson\thanks{Corresponding Author: sam.bilson@npl.co.uk}}
\author[1]{Maurice Cox}
\author[2,3]{Anna Pustogvar}
\author[1]{Andrew Thompson\thanks{Corresponding Author: andrew.thompson@npl.co.uk}}
\affil[1]{Department of Data Science, National Physical Laboratory, Teddington, UK}
\affil[2]{Department of Thermal \& Radiometric Metrology, National Physical Laboratory, Teddington, UK}
\affil[3]{School of Geography, Geology \& the Environment, University of Leicester, Leicester, UK}

\date{}

\begin{document}

\maketitle

\begin{sloppy}
\begin{abstract}
    Machine learning (ML) classification models are increasingly being used in a wide range of applications where it is important that predictions are accompanied by uncertainties, including in climate and earth observation, medical diagnosis and bioaerosol monitoring. The output of an ML classification model is a type of categorical variable known as a nominal property in the International Vocabulary of Metrology (VIM). However, concepts related to uncertainty evaluation for nominal properties are not defined in the VIM, nor is such evaluation addressed by the Guide to the Expression of Uncertainty in Measurement (GUM). In this paper we propose a metrological conceptual uncertainty evaluation framework for nominal properties. This framework is based on probability mass functions and summary statistics thereof, and it is applicable to ML classification. We also illustrate its use in the context of two applications that exemplify the issues and have significant societal impact, namely, climate and earth observation and medical diagnosis. Our framework would enable an extension of the GUM to uncertainty for nominal properties, which would make both applicable to ML classification models.
\end{abstract}
\end{sloppy}
\newpage
\section{Introduction}
\label{sec:Introduction}

\subsection{Machine learning classification from a metrology perspective}\label{subsec:ML}

The use of machine learning (ML) models offers the potential to bring great benefits to a wide range of societal challenges, for example in climate and earth observation~\cite{ESA2022, ESA2023, Skakun_2022,buters_automatic_2024,erb_real-time_2024}, health care~\cite{venton2021robustness,torres2020multi}, and advanced manufacturing~\cite{bilson2024machine}. ML models can be used to make sense of large data sets and learn complex relationships within them, and also to automate manual processes that require considerable time or expertise. 

Many of these applications involve the measurement of some signal on an instrument and inferring information of value to society from that signal. Traditionally, such problems would have involved developing a physical or empirical model that would relate the measurements to the information of interest. Where the information of interest is a quantity or set of quantities, the empirical model developed would form what is considered in metrology a \emph{measurement model}, providing the relationship between input quantities and a \emph{measurand}, the ``quantity intended to be measured''~\cite{JCGMVIM3}.

ML methods, and in particular methods of \emph{supervised learning}, are increasingly being used to establish (define and parametrise) measurement models. In supervised learning, data-driven models are built to infer predictions of the value of some (output) variable given knowledge of the values of other (input) variables. Typically a model \emph{type} is first selected, and then its parameters are optimised. An example of a type of model used in supervised learning is a neural network with a predefined architecture (see Section~\ref{sec:AF}). The phrase ``to learn the model'' is commonly used in ML to refer to the optimisation of the model parameters. 

The distinctive feature of supervised learning models is that their parameters are learned from training data, rather than purely using physical intuition or laws. Two types of supervised learning are often distinguished: \emph{regression} and \emph{classification}. The distinction is based on the type of variable that is being predicted: a regression model predicts a quantitative variable, while a classification model predicts a categorical variable. In Section~\ref{sec:CategoricalVar} we define the terms \emph{quantitative} and \emph{categorical} more precisely and provide examples of each.





Both regression and classification models have wide application, but the focus of this paper is on classification models. By a \emph{classification model} we mean a function~$f$, which takes some number of input variables and returns a categorical output variable. Formally speaking, we can denote a \emph{collection} of input variables by $X$. A collection is like a vector but can include both quantitative and categorical variables. The output categorical variable is denoted by $Y$, and can take one of $K$ values (often referred to in ML as classes) denoted by $\{c_1,\dots,c_K\}$. Following the convention of the Guide to the Expression of Uncertainty in Measurement (GUM)~\cite{jcgm-2023gum1}, we denote a value taken by any of these variables using lower case. Thus $x$ is an instance of $X$ and $y$ is an instance of $Y$ and these are related through 
\begin{equation}
    y=f(x),\quad y\in\mathcal{C}_K:=\{c_1,\dots,c_K\}.
    \label{y=f(X)}
\end{equation}
In reality, most classification models output a particular type of categorical variable known as a \emph{nominal} variable (sometimes referred to as an unordered categorical variable), or a \emph{nominal property} in the language of the International Vocabulary of Metrology (VIM)~\cite{JCGMVIM3}, in which the categories have no natural ordering. See Section~\ref{sec:CategoricalVar} for a more detailed discussion of variable types and how they are categorised in the VIM.

More specifically, an \emph{ML classification model} is one that has been learned from labelled training data, by which we mean $N$ samples (or observations) of paired data $(x_i,y_i),\,i=1,\dots,N$, consisting of observations of both the collection of input variables $X$ and the output variable $Y$. The supervised learning process consists of two stages: a training phase and a prediction phase. In the training phase, the model $f$ is learned from known paired inputs and outputs, which typically involves optimising the parameters of the model. In the prediction phase, the model is then evaluated upon new input variables. ML classification models, which are learned from training data, can be distinguished from \emph{rule-based} classifiers which use predetermined classification rules which are not dependent upon data.

\subsection{Uncertainty evaluation for machine learning classification}\label{sec:uncEval}

It is always important that an ML classification model is deployed in a trustworthy and quality-assured way. More specifically, ML classification models are currently being used in several applications in which it is crucial that predictions are accompanied by an evaluation of uncertainty. Examples include earth observation such as land cover classification, mapping of water, ice extent and cloud masking~\cite{ESA2022, ESA2023, Skakun_2022}, environmental monitoring such as bioaerosol classification~\cite{buters_automatic_2024,erb_real-time_2024}, and medical diagnosis such as the detection of heart arrythmias from electrocardiogram (ECG) and photoplethysmography (PPG) signals~\cite{venton2021robustness,torres2020multi}. See Section~\ref{sec:caseStudies} where we illustrate the conceptual contributions of this paper in the context of two case studies: land cover classification and atrial fibrillation detection. The importance of uncertainty evaluation for ML was highlighted in the Strategic Research Agenda developed by the European Metrology Network for Mathematics and Statistics (MATHMET)~\cite[Section 3]{mathmet2024SRA}.

In a metrology context, the outputs of an ML model may also be used to make inferences about other variables, that is, they can be used as input quantities for further measurement models or classification models. It follows that, as well as being important as an end in itself, uncertainty evaluation also enables the outputs of ML models to be used as part of an ongoing traceability chain, where the term \emph{traceability chain} is used to include both traditional metrological traceability chains, where artefacts or reference materials are passed between facilities, and situations where data are passed through multiple processing steps (e.g.~\cite{woolliams2025metrological}). In Section~\ref{sec:pmfProp}, we provide an example of how the outputs of an ML classification model, along with their accompanying uncertainties, can be included as inputs to other models.

\subsubsection{The uncertainty evaluation task for machine learning classification models}
\label{The uncertainty evaluation task for ML classification models}

For classification models, a common way to express the confidence of a prediction is to assign a probability to each of the $K$ possible classes rather than simply returning a single predicted class~\cite[Definition 2.9]{JCGMVIM3}. Assigning a probability to each class of a nominal property is equivalent to obtaining the probability mass function (PMF) for the classification. In Section~\ref{sec:UQ} we develop the observation that the role of the PMF here is the natural counterpart of the role of the probability density function (PDF) in the measurement of a continuous quantitative variable (continuous quantity). We also show in Section~\ref{sec:UQ} how the PMF of a categorical variable can be used as an input into other models, allowing the propagation of uncertainty through multistage measurement models involving nominal properties.

Section~\ref{subsec:ML} described how ML classification is a type of supervised learning where a model is learned that relates outputs to inputs, and where an implementation of ML classification involves a training phase followed by a prediction phase. In ML classification models, contributions to the overall uncertainty of a prediction are introduced in both the training and prediction phases, which is in contrast to rule-based classifiers in which there is no training phase and where the classification rule is predetermined, for example tolerance thresholds in conformity assessment (see also Section~\ref{sec:metFrame}). 

A crucial assumption in supervised learning is that the model learned on training data is applicable to input data for which predictions are desired. One way to statistically characterise this assumption is to say that the input data (for both training and prediction) are independent and identically distributed (i.i.d.). In practice, this assumption must be validated through appropriate experimentation. For example the i.i.d. assumption can be validated using a suitable hypothesis test. Where the assumption is violated, new input data for which model predictions are required is said to have undergone \emph{distribution shift}; see for example~\cite[Chapter 8]{huyen2022designing}. Uncertainty evaluation methods for supervised learning predictions typically assume the absence of distribution shift, and such methods are only reliable to the extent that this assumption is not violated.

\begin{figure}[ht]
\centering
    \includegraphics[width=0.9\textwidth]{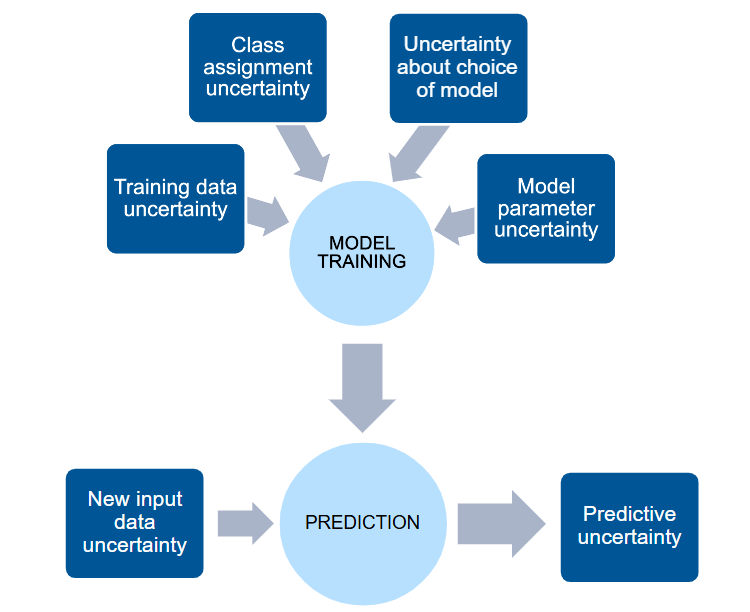}
    \caption{Factors contributing to uncertainty in ML prediction.}
    \label{fig:uncertainty_sources}
\end{figure}

We next outline some of the factors contributing to prediction uncertainty throughout the ML pipeline. These contributory factors are also illustrated in Figure~\ref{fig:uncertainty_sources}.

\begin{enumerate}
\item \label{training data uncertainty} \textbf{Training data uncertainty.} In the training phase, one contribution to prediction uncertainty is uncertainty in the training data itself. For example, when classifying landcover based upon surface reflectance measurements (see our case study in Section~\ref{sec:landcover}), there is uncertainty in the surface reflectance measurements due to imperfect knowledge of aerosol optical depth (input variables), and the landcover data is also subject to labelling errors (output data). These training data uncertainties affect the training of the model.
\item \label{ill-posed}
\textbf{Class assignment uncertainty.} In classification tasks for which ML might be used, there is typically no theoretical justification for the choice of input variables, and therefore no reason to assume that observations of the input variables are sufficient to determine a classification unambiguously. Such tasks are in this sense ill-posed: a given observation for the input variables could occur for data drawn from more than one class. In such cases, there is a limitation, inherent in the modelling task, to determining the class to which a data sample belongs. This limitation is itself dependent upon the choice of input variables. \emph{Feature selection} refers to the common practice in ML of choosing a subset of available input variables, which can help to remove redundant/unnecessary input variables. However, feature selection cannot eliminate class assignment uncertainty if the original selection of variables is insufficient. 

\item \label{choice of model} \textbf{Uncertainty about choice of model.} The supervised learning approach relies upon the selection of a certain type of classification model (see Section~\ref{subsec:ML}). The model type is typically selected using empirical testing, and this process introduces uncertainty. Bayesian model averaging is one method that has been proposed to capture uncertainty due to choice of model~\cite{hoeting1999bayesian}.


    
\item \label{parameter uncertainty} \textbf{Model parameter uncertainty.} ML models need to be optimised using the training data, for example, optimising the values of its parameters. There is, therefore, also uncertainty related to the estimates of the ML model parameters. Some of this uncertainty may be due to the challenges of optimising a model with multiple local optima. The ability to optimise an ML model also depends upon having training data that samples the region of data space that is of interest sufficiently densely. Training data requirements vary depending upon the complexity of the classification task and the chosen ML model.

\item \label{prediction data uncertainty} \textbf{New input data uncertainty.} In the prediction phase, additional uncertainty is introduced through measurement uncertainties in previously unused (non-training) input data to which the model is being applied.
\end{enumerate}

Uncertainty source (\ref{ill-posed}) can be understood as inherent to the learning task. 
Uncertainty sources (\ref{training data uncertainty}) and (\ref{prediction data uncertainty}) constitute properties of the measuring system used to obtain the observational data. 
Uncertainty sources (\ref{choice of model}) and (\ref{parameter uncertainty}) are both a property of the modelling approach (though modelling approaches are often constrained by the task). 

Many modern ML classifiers were originally developed with the aim of identifying the (single) correct class. However, the need to go beyond deterministic classifiers and evaluate uncertainty has been recognised in the ML community; see for example~\cite{hullermeier2021aleatoric}. Many, though not all, ML classifiers are probabilistic, in the sense that they return class probabilities~\cite{murphy2012machine}. Methods differ according to which sources of uncertainty are captured and whether a Bayesian or frequentist approach is adopted.

In Section~\ref{sec:caseStudies} we present two case studies based on societal applications that rely on a metrological analysis of probabilistic ML classifiers.

In our first case study (Section~\ref{sec:landcover}), a Bayesian approach is taken using a Quadratic Discriminant Analysis (QDA) model~\cite{bilson2025uncertaintyawarebayesianmachinelearning}. This method explicitly models four of the five sources of uncertainty described above and illustrated in Figure~\ref{fig:uncertainty_sources}. The method falls into the general category of \emph{generative modelling} in which the input variables in each class are assumed to follow some statistical distribution, which in the case of QDA is a multivariate Gaussian distribution. Other examples of generative classifiers include Linear Discriminant Analysis and Na\"ive Bayes; see~\cite{murphy2012machine} for more details.

Many modern ML classifiers are instead examples of \emph{discriminative modelling} in which classification is achieved without making statistical assumptions about the data. In our second case study (Section~\ref{sec:AF}), a discriminative modelling approach based on neural network models known as \emph{Monte Carlo Dropout} is used~\cite{gal2016dropout,kendall2017uncertainties}. This method captures uncertainty arising from both the data and the model, but it does not explicitly model the various sources of uncertainty in Figure~\ref{fig:uncertainty_sources}. The method does, however, have the advantage of being applicable to high-complexity neural networks, which are known to be especially able to capture complex non-linear relationships. Other examples of discriminative classifiers include logistic regression, support vector machines and tree-based methods such as random forests; see~\cite{Hastie2008} for more details.

\subsubsection{Related metrological frameworks}\label{sec:metFrame}

Evaluation of the uncertainty of quantitative outputs from measurement models is addressed by the GUM suite of documents~\cite{jcgm-2023gum1,JCGMGUM,JCGM101,JCGM102} and the related VIM vocabulary~\cite{JCGMVIM3}. Central to the GUM is the concept of the propagation of measurement uncertainty through measurement models whose parameters have been determined. This scenario would correspond to a trained ML model in which estimates of the parameters are used.  The GUM can also treat the parameters that define the measurement model as input quantities in their own right, propagating uncertainty (and where there are multiple such parameters, their covariance) as with other input quantities. 



The GUM framework is explained for models obtained through regression of parametric models (empirical measurement models) in GUM-6~\cite{jcgm-2020gum6}. Uncertainty associated with model parameter estimates is addressed in~\cite[clause 7]{jcgm-2020gum6}. Uncertainty concerning the choice of model is discussed briefly in~\cite[clause 8.5]{jcgm-2020gum6}. The inclusion of effects due to measurement uncertainties in either the input or output variables is addressed in~\cite[clauses 9 and 10]{jcgm-2020gum6}.

The frameworks developed in the GUM suite of documents can thus, in principle, be applied to any ML regression model. However, the GUM documents mentioned above do not address classification models since the concept of measurement uncertainty evaluation has been developed in the metrology community (and in the GUM and VIM documents) for quantitative variables, or \emph{quantities} in the language of the VIM and the GUM. The output of a classification model, on the other hand, is not a quantitative variable, but rather an unordered categorical variable, or a nominal property.

\begin{figure}[ht]
\centering
    \includegraphics[width=\textwidth]{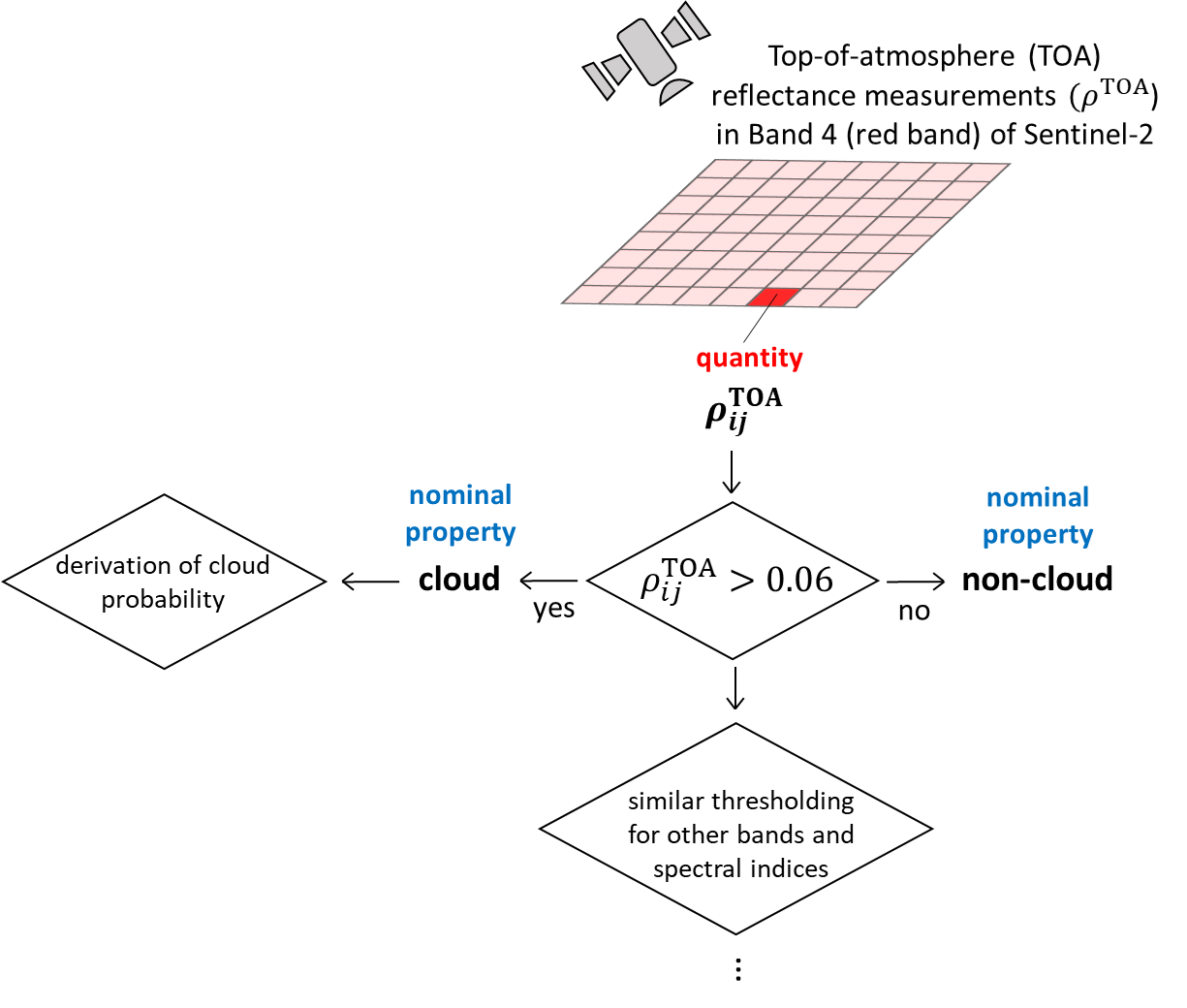}
    \caption{Cloud masking of Sentinel-2 satellite imagery.}
    \label{fig:cloud_masking}
\end{figure}

A certain kind of classification model is addressed in the fourth document of the GUM suite dedicated to conformity assessment~\cite{JCGM106-2012}. In conformity assessment, the result of a measurement, which represents a quantity (e.g., mass concentration of mercury in a sample of industrial water), is converted into a nominal variable, which has two categories: \emph{conforming} and \emph{non-conforming}. The conversion is implemented by identifying whether the result of a measurement falls into an interval of permissible values set by one or two tolerance limits. Such conversion essentially represents a binary classification task. Binary classification performed by applying a threshold to a quantity (similar to conformity testing) has many applications. For example, in earth observation, operational cloud masking of satellite imagery has been historically performed by applying empirically derived thresholds to top-of-atmosphere reflectance measurements, e.g., see~\cite{Louis2021} (Figure~\ref{fig:cloud_masking}). However, it is important to note that with the recent progress of ML learning, and specifically deep learning, more efforts have been made in developing more advanced cloud masking algorithms (e.g,~\cite{LopezPuigdollers2021}).

Similarly to a probabilistic ML classifier, in conformity assessment the assigned category is provided with a probability value, the \emph{conformance probability}. However, in conformity assessment the classifier is rule-based, there is no training phase, and therefore uncertainty related to a training phase does not need to be taken into account. For these reasons, the framework in~\cite{JCGM106-2012} cannot be used for a comprehensive uncertainty evaluation of ML classifiers, which rely on training.

In a standard conformity assessment a threshold is fixed (e.g., a certain critical mass concentration of mercury based on which a water sample is considered to be conforming or non-conforming). In contrast, in a binary classification performed alike conformity assessment (e.g, cloud masking described above), a threshold is often empirically derived (e.g., based on expert knowledge of top-of-atmosphere reflectance of a cloud vs. non-cloud). In the latter case, there would be an additional uncertainty source associated with the choice of the threshold, which will affect the classification results, and this uncertainty source is not considered in a standard conformity assessment~\cite{JCGM106-2012}.

\subsection{Summary of contributions}

\begin{itemize}
\item We propose a metrological conceptual uncertainty evaluation framework for nominal properties which is applicable to ML classification. Our framework is inspired by existing vocabularies and frameworks from the fields of qualitative chemical analysis~\cite{bettencourt_da_silva_eurachemcitac_2021} and clinical laboratory sciences~\cite{nordin_vocabulary_2018}. It is based upon the determination of the PMF of the nominal variable returned by the ML classifier, and the expression of uncertainty in terms of summary statistics derived from it. We review some possible summary statistics, discussing their benefits and drawbacks. The framework represents a natural analogue to the determination of the PDF for a quantitative variable and the derivation from it of summary statistics such as standard uncertainty. We also show how conditional probability can be used to enable uncertainty evaluation for measurement models that have quantitative and categorical variables as inputs.





\item We illustrate our conceptual framework by applying it in the context of two case studies where a metrological approach adds value. The first is the use of ML for classification of land cover based upon surface reflectance measurements made by a spaceborne optical sensor. The second is the use of ML for detection of atrial fibrillation based upon photoplethysmography (PPG) signals used in wearable devices.
\end{itemize}

\subsection{Outline of contents}

In Section~\ref{sec:CategoricalVar}, we give background on nominal properties, and we describe how such properties are addressed in metrology. In Section~\ref{sec:UQ}, we present our framework for uncertainty evaluation of nominal properties based on discrete probability distributions and summary statistics. We also address the propagation of uncertainty through multistage measurement models. We present our two ML case studies illustrating the framework in Section~\ref{sec:caseStudies}, before concluding in Section~\ref{sec:conclusion}.

\pagebreak
\section{Nominal variables}
\label{sec:CategoricalVar}

\subsection{Nominal versus other variables}
Classification tasks deal with nominal variables. To explain what nominal variables are, we will use Stevens' theory of scales of measurement~\cite{stevens_theory_1946}. For a deeper review of Stevens’ theory and its relevance to metrology we refer the reader to~\cite{white_meaning_2011}. Stevens' theory distinguishes four types of scales of measurement: \emph{nominal}, \emph{ordinal}, \emph{interval} and \emph{ratio} (see Table~\ref{table:Stevens theory}). A nominal scale has multiple categories (or classes) with no natural ordering, for example a biological species or a blood group. Although these categories can be represented by numbers, the numbers do not have any quantitative meaning. Properties measured on a nominal scale can only be compared based on their equivalence, i.e., same or different.

An ordinal scale also has multiple categories, but with a natural ordering, for example wind force on the Beaufort scale or stages of cancer. Properties measured on an ordinal scale can be compared based on their equality and their order, e.g., higher/lower, stronger/weaker. Distances between categories, however, are not necessarily meaningful. For example, the difference in the severity of cancer stages I and II is not the same as the difference in the severity of cancer stages III and IV~\cite{ISO33406}.

Unlike for an ordinal scale, for an interval scale the differences are meaningful. Examples of properties measured on an interval scale include longitude and latitude, year, etc. A distinctive feature of this scale is that it does not have a meaningful (`true') zero, which is chosen arbitrarily.

Properties measured on a ratio scale do have a `true' zero and represent the most encountered properties in physics, for example length, mass, and electrical power.

\begin{table}[ht]
\caption{Stevens' scales of measurement}
\label{table:Stevens theory}
\begin{center}
\begingroup
\setlength{\tabcolsep}{6pt} 
\renewcommand{\arraystretch}{1.25} 
\begin{tabular}{>{\raggedright}lm{6cm}>{\raggedright}m{3.2cm}>{\raggedright\arraybackslash}m{2.3cm}}
 \toprule
 Scale & Examples & Central tendency statistic & Dispersion statistic  \\
 \midrule\midrule
 nominal & biological species, blood group & mode & see Section 3\\
 \midrule
 ordinal & wind force on the Beaufort scale, stage of cancer & median, mode & interquartile range (IQR) \\
 \midrule
 interval & longitude and latitude, year  & mean, median, mode & standard deviation, IQR \\
 \midrule
 ratio & length, mass & mean, median, mode & standard deviation, IQR \\
\bottomrule
\end{tabular}
\endgroup
\end{center}
\end{table}

One of the important aspects of Stevens' theory is that different types of scales have different statistics representing central tendency and dispersion. From a statistical point of view, a nominal scale represents the lowest (the most primitive) scale, while a ratio scale represents the highest scale. Statistics that apply to the lower scales also apply to the higher scales, but not the other way around~\cite{Agresti2002}. The central tendency of properties measured on a nominal scale can only be expressed by the mode (although strictly speaking, the mode is representative of the most frequent class, not the central class). The central tendency of properties measured on an ordinal scale can be expressed by the mode and the median. The notions of central tendency and dispersion are important for understanding how to express uncertainties. In Section~\ref{sec:UQ}, we discuss expressions of uncertainty for nominal variables.

The nomenclature for nominal variables varies across scientific communities. In the mathematical sciences, nominal and ordinal variables are often merged into a single group, \emph{categorical variables}~\cite{Agresti2002,Hastie2008}, with nominal variables sometimes referred to as \emph{unordered categorical variables} and ordinal variables referred to as \emph{ordered categorical variables}~\cite{Hastie2008}. In addition, nominal variables are sometimes called \emph{qualitative variables}, and are distinguished from \emph{quantitative variables} (variables measured on interval and ratio scales), with some ambiguity around the classification of ordinal variables~\cite{Agresti2002}. However, in some international standards, the term \emph{qualitative variable} (or \emph{qualitative property}) is used as an umbrella term for both nominal and ordinal variables~\cite{ISO33406}.

In this paper, we use the term \emph{categorical variable} as an umbrella term for nominal and ordinal variables. We do not use the term \emph{qualitative variable} to avoid confusion because, as mentioned above, there is disagreement about whether ordinal variables are included~\cite{Agresti2002,ISO33406}. We also use interchangeably the term \emph{variable} (more common in the language of data science) and the term \emph{property} (more common in the language of metrology).

\subsection{Nominal variables in metrology}\label{sec:IntStand}

The authoritative metrological documents used to focus solely on quantities (quantitative properties). This focus changed about a decade ago, when definitions of nominal properties and ordinal quantities were introduced in the 3$^\text{rd}$ edition of the VIM, the VIM3~\cite[Definitions 1.30 and 1.26]{JCGMVIM3}. In the preparation of the 4$^\text{th}$ edition of the VIM (the VIM4), the Joint Committee for Guides in Metrology (JCGM) is holding a discussion on including an entire chapter on nominal properties, motivated by the growing interest in a metrologically rigorous handling of these properties.

Unlike the VIM, the GUM suite of documents is entirely focused on quantities, with
the only exception being the fourth document of the GUM suite dedicated to conformity
assessment~\cite{JCGM106-2012}, previously mentioned in Section~\ref{sec:Introduction}.

Although the VIM and the GUM currently provide negligible coverage of nominal variables, these variables have been addressed in a number of standards and recommendations produced by international bodies working in various areas of science and industry, where nominal variables are frequently used. For example, the International Federation of Clinical Chemistry and Laboratory Medicine (IFCC) and International Union of Pure and Applied Chemistry (IUPAC) developed ``Vocabulary on nominal property, examination, and related concepts for clinical laboratory sciences (IFCC-IUPAC Recommendations 2017)"~\cite{nordin_vocabulary_2018}. Also, nominal variables and approaches to the evaluation of uncertainties associated with such variables were discussed in ``Assessment of performance and uncertainty in qualitative chemical analysis (EURACHEM/CITAC Guide)"~\cite{bettencourt_da_silva_eurachemcitac_2021}. Recommendations on handling nominal variables specifically related to reference materials, which have nominal properties, were recently systemised in the International Standard ISO 33406:2024 ``Approaches for the production of reference materials with qualitative properties"~\cite{ISO33406}.
Discussion on nominal variables and their uncertainty evaluation can also be found in publications by national metrology institutes (e.g.,~\cite{possolo_simple_2015}), as well as in publications by the members of JCGM Working Group 2 ~\cite{mari_foundations_2020}. It is important to note that although many of the above references address evaluation and expression of uncertainties associated with nominal variables, no agreement has yet been reached on this topic.

\pagebreak
\section{Uncertainty evaluation for nominal properties}\label{sec:UQ}
In this section, we outline approaches for uncertainty evaluation of nominal properties and further propagation through measurement models. In Section~\ref{sec:pd}, we define PDFs and PMFs, illustrating that PMFs are the natural probability distribution for nominal properties. In Section~\ref{sec:exUnc}, we describe various statistics that could be used to express uncertainty of nominal variables, highlighting their strengths and weaknesses and the extent of their departure from accepted notions of uncertainty in the metrology community. In Section~\ref{sec:pmfProp}, we show, with use of the PMF, how uncertainties of nominal properties can be propagated through measurement models having categorical inputs. 

\subsection{Probability distributions}\label{sec:pd}
As stated in the GUM~\cite{jcgm-2023gum1} and other metrology-related publications~\cite{possolo_simple_2015}, the uncertainty associated with any variable is fully expressed by its probability distribution, which is the distribution around a measured value (often taken as the mean of the distribution) in which the `true' value is expected to lie. This distribution can either be empirical (i.e., based upon the frequency of occurrence of the variable), or have an assumed form (such as Gaussian). For continuous quantitative variables $\mathbf{x}$ that can take values in the set $\mathcal{X}\subset\mathbb{R}^D$, where $D$ is the dimension of the continuous input space, the probability distribution is defined in terms of a PDF~\cite{evans2004probability}\cite[Definition 2.26]{ISO3534}, which is a map or function 
\begin{eqnarray}
    \label{eq:PDF}
    p:\mathcal{X}&\rightarrow&\mathbb{R}^+\nonumber, \\
    \mathbf{x}&\mapsto&p(\mathbf{x}),
\end{eqnarray}
where
\begin{equation}
\label{eq:PDFsum}
    \int_{\mathcal{X}}p(\mathbf{x})\,\mathrm{d}\mathbf{x}=1.
\end{equation}
Condition~\eqref{eq:PDFsum} ensures that expression~\eqref{eq:PDF} is properly normalised and can act as a PDF.

Similarly, a nominal variable can take values in the set $\mathcal{C}_K=  \{c_1, \dots, c_K\}$, where $K$ is the number of possible classes (or categories). Thus, the probability distribution is defined in terms of the PMF~\cite{murphy2012machine}\cite[Definition 2.24]{ISO3534}:
\begin{eqnarray}
    \label{eq:PMF}
    p:\mathcal{C}_K&\rightarrow&[0,1], \nonumber\\
    c_k&\mapsto&p_k:=p(c_k),
\end{eqnarray}
where
\begin{equation}
    \label{eq:PMFsum}
    \sum_{k=1}^Kp_k=1.
\end{equation}
Condition~\eqref{eq:PMFsum} ensures the expression~\eqref{eq:PMF} obeys the correct probability sum rule and can act as a PMF.

The PMF of a nominal property can also be fully described in terms of its \emph{probability vector}
\begin{equation}
    \mathbf{p}=[p_1, \dots, p_K] \in [0,1]^K,
\end{equation}
where each element of the array can take values on the interval $[0,1]$, and condition~\eqref{eq:PMFsum} also holds.

Given the analogy with PDF for quantitative variables, the PMF is the natural candidate for expressing uncertainty in nominal properties and has been motivated in metrology-related publications such as~\cite{mari_foundations_2020,possolo_brief_2024}, as well as the recent international standard~\cite{ISO33406} introduced in Section~\ref{sec:IntStand}.

\subsection{Expressions of uncertainty}\label{sec:exUnc}

 Expressions of uncertainty for quantitative variables such as the standard deviation are well known in the metrology community, and are used in the GUM~\cite{jcgm-2023gum1}. The standard deviation is one example of a summary statistic that expresses the uncertainty of a quantitative variable in terms of a single positive real number. More formally, given a quantitative variable $\mathbf{X}$ with PDF $\mathbf{p}(\mathbf{x})$, a statistic $u$ for uncertainty is evaluated using a function $F$ such that 
\begin{equation}
    u=F(\mathbf{p}(\mathbf{x})),\quad u\in[0,\infty).
\end{equation}

The definition of uncertainty according to the VIM~\cite[2.26]{JCGMVIM3} is, a ``non-negative parameter characterizing the dispersion of the quantity values being attributed to a measurand, based on the information used.'' A statistic of dispersion is typically assumed to involve some notion of distance, and it is has been questioned whether such a notion exists for nominal properties~\cite{mari_foundations_2020}. The PMF of a nominal property can be interpreted in terms of expected distance with respect to a simple metric. Using the Hamming distance on a single categorical variable~\cite[Chapter 25]{murphy2012machine}
\begin{equation}
    \label{eq:distDef}
   s_{jk}=\begin{cases}
        0 & c_j=c_k, \\
        1 & \text{otherwise},
    \end{cases}\quad j,k\in\{1,\dots,K\},
    \end{equation} 
the expected distance to class $c_k$ of a nominal property following the distribution given by $\mathbf{p}$ is given by
\begin{equation}\sum_{j=1}^K p_j s_{jk}=\sum_{j\neq k} p_j=1-p_k.
\end{equation}

Various so-called \emph{statistics of variation}~\cite{kvalseth_variation_2011} have been proposed for nominal properties, and we present an overview of some of these in Section~\ref{sec:varStat}. Given our observation in the previous paragraph that all summary statistics of a PMF have a distance interpretation, we will refer to them throughout as statistics of dispersion.

The VIM also defines a notion of \emph{standard measurement uncertainty} for quantities, namely the standard deviation, which captures dispersion about a central tendency, namely the mean~\cite[(2.30)]{JCGMVIM3}. For nominal properties, the only statistic of central tendency is the mode. Similarly to the notion of distance in (\ref{eq:distDef}), we can also define a meaningful notion of \emph{distance from the modal class} (or classes in the multimodal case) for nominal properties (see Appendix~\ref{app}). Equipped with this notion of distance, some of the statistics we present in Section~\ref{sec:varStat} can indeed be viewed as statistics of dispersion about the modal class(es). Writing $\widehat{p}=\max(\mathbf{p})$ for the probability assigned to the modal class(es), also known as the \emph{modal probability}, those statistics that can be expressed as a function of $\widehat{p}$ alone can be interpreted as dispersion about central tendency. 

That said, some of the statistics we present in Section~\ref{sec:varStat} are not in general statistics of dispersion from a central tendency. Such statistics do not, therefore, represent a counterpart to the notion of standard measurement uncertainty for quantities. 

Such a distinction is only relevant for multi-class nominal variables ($K>2$). For binary nominal variables ($K=2$), all statistics of dispersion (from the mode or otherwise) can be expressed as a function of $\widehat{p}$ alone, and so capture dispersion about central tendency. For each of the statistics of dispersion defined in Section~\ref{sec:varStat}, we state their explicit reduction to a function of $\widehat{p}$ alone in the binary case. 



\subsubsection{Uncertainty and confidence}\label{sec:uncConf}

A distinction is made in the literature e.g.,~\cite{mari_foundations_2020,shirmohammadi_measurement_2024}, between the \emph{confidence} in the prediction of an ML classification model, and the predictive \emph{uncertainty}. Given a probability vector $\mathbf{p} = [p_1, ,\dots, p_K]$ of an ML classification model $y = f(x),\, y \in \mathcal{C}_K$, the classification of the output is typically given by the mode of the PMF, defined as the \emph{modal class} $\widehat{c}$. The confidence in the prediction can then be expressed as modal probability $\widehat{p}$, which is the probability that $y$ is the modal class. Note that this notion of confidence is distinct from a \emph{confidence interval} which is one way of expressing uncertainty for quantitative variables in the GUM and the VIM.

Using the modal probability as a form of uncertainty expression has advantages in the sense that it is directly associated with the model prediction, and as such aligns with the language for expressing uncertainty of quantitative variables~\cite{jcgm-2023gum1}. 
For example, the output of a classification model $\widehat{y} = f(x)$ can be written as the pair $(\widehat{c},\widehat{p})$, where $\widehat{p}$ is the probability of the prediction $\widehat{y} = \widehat{c}$. This language can be extended~\cite{mari_foundations_2020} to include multiple classes, i.e., the ML classification model predicts $\widehat{y}=\{c_1,c_2\}$ with probability $p=p_1+p_2$. However, confidence is not the same as uncertainty. A probability equal to $1$ would express complete certainty in the output of a classification model, thus the uncertainty should be $0$. It is possible, though, to define a statistic of dispersion about a central tendency which is a linear transformation of confidence (see Section~\ref{sec:varStat}). Thus, even though uncertainty and confidence are not the same, they are intimately related.

\subsubsection{Properties of an uncertainty statistic}
\label{sec:uncProp}
In general, a statistic $F$ expressing the uncertainty associated with a nominal variable $y$ with PMF $\mathbf{p}$, should satisfy the following properties (adapted from~\cite{wilcox_indices_1973,kvalseth_variation_2011}):
\begin{enumerate}
    \item $F$ gives a non-negative real scalar value: $F(\mathbf{p}) \in [0, \infty)$.
    \item $F$ gives $0$ if and only if there is no uncertainty. That is, $F(\mathbf{p}^1) = 0$ if and only if the elements of $\mathbf{p}^1$ take the form
    \begin{equation}
        p^1_k := \begin{cases}1 & c_k = \widehat{c}, \\ 0 & \text{otherwise,}\end{cases}\quad k=1,\dots,K.
        \end{equation}
    \item $F$ is maximum when there is maximal uncertainty. This is when all probabilities are equal, or equivalently when all classes could be the mode: 
    \begin{equation}
        F(\mathbf{p})\leq F(\mathbf{p}_\text{unif}),
    \end{equation} 
    where $(p_\text{unif})_k=1/K,\,k=1,\dots,K$ is the categorical uniform distribution.
\end{enumerate}
A normalised statistic $F_{\mathrm{norm}}$ can then be defined for nominal properties:
\begin{equation}
    F_{\mathrm{norm}}(\mathbf{p}) = \dfrac{F(\mathbf{p})}{F(\mathbf{p}_\text{unif})} \in [0, 1],
\end{equation}
where $F_{\mathrm{norm}}(\mathbf{p}_\text{unif})=1$ represents maximal uncertainty.

In summary, a (normalised) statistic $F$ expressing the uncertainty associated with the output probabilities of an ML classification model is minimum (zero) when the model predicts  a single class with $100\,\%$ probability, and is maximum (one) when the model predicts equal probabilities to all classes.

In the following analysis, we will consider only normalised statistics that satisfy the above properties, for a fair comparison between expressions of uncertainty (see Section~\ref{sec:compStat}). As such, we do not consider statistics that quantify different aspects of uncertainty, such as the expected difference of information to a reference class~\cite{bogaert_information-based_2017}, which has been gaining popularity in land cover classification (see case study in Section~\ref{sec:landcover}). This statistic fails on property 3, since the statistic is minimum for a uniform distribution, whereas it should be maximum if it is to represent uncertainty.

\subsubsection{Statistics of dispersion}\label{sec:varStat}

In this section, we give an overview of some of the statistics of dispersion that have been proposed.
\vspace{5mm}
\par\textbf{Wilcox's variation ratio}

Wilcox's variation ratio (WVR)~\cite{wilcox_indices_1973}, also known as the mean deviation from the mode, is a natural extension of the concept of confidence (see Section~\ref{sec:uncConf}) in a classification model prediction to an expression of uncertainty, defined by
\begin{equation}
\label{eq:WVR}
    u_\mathrm{WVR}(\mathbf{p})=1-\dfrac{K\widehat{p}-1}{K-1},
\end{equation}
For unimodal distributions only, the WVR can be expressed in terms of the expected distance from the mode. Thus the WVR is analogous to the standard deviation for continuous variables. See Appendix~\ref{app} for a derivation of WVR in terms of expected distance from the mode.

For binary classification ($K=2$), expression~\eqref{eq:WVR} reduces to
\begin{equation}
    \label{eq:WVRbin}
    u_\mathrm{WVR}(\mathbf{p})=2(1-\widehat{p}).
\end{equation}
The WVR expresses the remaining uncertainty after taking into account the modal probability $\widehat{p}$ of the classification result. However, it does not take into account the entire PMF, and and in particular variation in probabilities of the remaining classes. This issue is only relevant in the case of non-binary classification problems, such as the one described in Section~\ref{sec:landcover}.
\vspace{5mm}
\par\textbf{Universal variation ratio}

The universal variation ratio (UVR)~\cite{rubia_measures_2024} is an extension of the WVR that explicitly takes account of multimodal distributions, defined as
\begin{equation}
\label{eq:UVR}
    u_\mathrm{UVR}(\mathbf{p})=\dfrac{K^2}{K^2-1}\left(1-\dfrac{\widehat{p}}{m}\right),
\end{equation}
where $m$ is the number of modes. This extension allows expression of the UVR in terms of the expected distance from the mode in the multimodal case (see Appendix~\ref{app} for a derivation), which cannot be achieved for the WVR. 

By explicitly including the number of modes, the UVR is discontinuous as $\widehat{p}$ varies. This becomes clear in binary classification, where $m=1$ in all cases except for a uniform distribution when $m=2$. Therefore, in binary classification expression~\eqref{eq:UVR} reduces to
\begin{equation}
    u_\mathrm{UVR}(\mathbf{p})=\begin{cases}
        \frac{4}{3}(1-\widehat{p}) & \widehat{p}>0.5,\\
        1 & \widehat{p}=0.5.
    \end{cases}
\end{equation}
It can be shown~\cite{rubia_measures_2024} that $u_\mathrm{UVR}(\mathbf{p})\leq u_\mathrm{WVR}(\mathbf{p})$ for all unimodal distributions.
\vspace{5mm}
\par\textbf{Standard deviation from the mode}

Both the WVR and UVR take account of the modal probability $\widehat{p}$ only, but not the distribution among the remaining classes, which might be important to consider. Motivated by finding a statistic based on the deviation from the mode, which takes account of all values in the PMF, Kv\aa lseth~\cite{kvalseth_measuring_1988} defined the standard deviation from the mode (SDM) for nominal variables. In normalized form this standard deviation is
\begin{equation}
\label{eq:SDM}
    u_\mathrm{SDM}(\mathbf{p})=1-\sqrt{\dfrac{\sum_{k=1}^K(\widehat{p}-p_k)^2}{K-1}}.
\end{equation}
For binary classification, expression~\eqref{eq:SDM} reduces to (see Appendix~\ref{app2}):
\begin{equation}
    \label{eq:SDMbin}
    u_\mathrm{SDM}(\mathbf{p})=2(1-\widehat{p}),
\end{equation}
and so in the binary case WVR and SDM are equivalent.

The SDM has the property that $u_\mathrm{SDM}(\mathbf{p})\leq u_\mathrm{WVR}(\mathbf{p})$ for all unimodal distributions.

One advantage of using the SDM is that for a large number $n$ of samples used to estimate the output PMF, such as the output trees of random forests, the SDM is normally distributed in the unimodal case~\cite{kvalseth_measuring_1988} $\widehat{u}_\mathrm{SDM}\sim\mathcal{N}(u_\mathrm{SDM}(\mathbf{p}),\sigma_\mathrm{SDM}^2(\mathbf{p}))$ with
\begin{equation}
    \sigma_\mathrm{SDM}^2(\mathbf{p})=\dfrac{\widehat{p}(1-K\widehat{p})^2+\sum_{k=1}^Kp_k(\widehat{p}-p_k)^2}{n(K-1)^2(1-u_\mathrm{SDM}(\mathbf{p}))^2}-\dfrac{(1-u_\mathrm{SDM}(\mathbf{p}))^2}{n}.
\end{equation}
This result allows the use of confidence intervals for the statistic. 

\vspace{5mm}
\par\textbf{Entropy}

The information (or Shannon) entropy (not to be confused with thermodynamic entropy) is an expression of information content, originally motivated to describe the theory of signal communication~\cite{shannon_mathematical_1948}.

The normalised version (or relative probability entropy) is defined for nominal properties as
\begin{equation}
\label{eq:ShEnt}
    H(\mathbf{p})=\mathbb{E}[-\log_K\mathbf{p}]=-\sum_{k=1}^Kp_k\log_Kp_k.
\end{equation}
In the binary classification case, expression~\eqref{eq:ShEnt} reduces to the original Shannon entropy in bits
\begin{equation}
    H(\mathbf{p})=-\widehat{p}\log_2(\widehat{p})-(1-\widehat{p})\log_2(1-\widehat{p}).
\end{equation}
Information entropy is commonly used in the existing literature as a statistic for expressing the uncertainty in nominal variables~\cite{mari_toward_2017,wilcox_indices_1973,possolo_statistical_2014}, as well as the recent international standard~\cite{ISO33406}. It is also the most commonly used statistic in the ML community for summarising uncertainty in predictions of classification models~\cite{kendall2017uncertainties,shirmohammadi_measurement_2024,hullermeier2021aleatoric}. The principle of maximum entropy (uncertainty) is also used in Bayesian inference (see Section~\ref{sec:BayesUnc}) for selection of prior probabilities~\cite{jaynes_prior_1968}.

Mari~\cite{mari_foundations_2020} suggested an extension to information entropy given (in normalised form) by
\begin{equation}
\label{eq:modEnt}
    H^*(\mathbf{p})=\dfrac{K^{H(\mathbf{p})}-1}{K-1}=\dfrac{\prod_{k=1}^Kp_k^{-p_k}-1}{K-1}.
\end{equation}
Such a transformation can be performed to any expression of uncertainty $F(\mathbf{p})$ for nominal properties to produce a new statistic $F^*(\mathbf{p})=(K^{F(\mathbf{p})}-1)/(K-1)$, which also satisfies the properties in Section~\ref{sec:uncProp}. This transformation has the property $F^*(\mathbf{p})\leq F(\mathbf{p})\,\forall\,\mathbf{p}$, and so will lower the variation in the values of the new statistic (see Tables~\ref{tab:LCbw2020},~\ref{tab:LCbw2021} and~\ref{tab:AFbw}). This allows the statistic to be more robust against small perturbations of the PMF.

One alternative variant on the information entropy is called the $\alpha$-quadratic entropy~\cite{pal_measuring_1994}, defined as
\begin{equation}
    \label{eq:aqEntropy}
    H_\alpha(\mathbf{p})=\frac{K^{2\alpha-1}}{(K-1)^\alpha}\sum_{k=1}^Kp_k^\alpha(1-p_k)^\alpha,\quad\alpha\in(0,1].
\end{equation}
If $\alpha$ is close to zero, this statistic is not very sensitive to changes in $\mathbf{p}$, but for $\alpha$ close to one, this statistic has higher sensitivity~\cite{low_impact_2013}, and has been used in evaluating uncertainties in land cover classification models~\cite{low_impact_2013,bogaert_information-based_2017}. 

Both of the statistics given in (\ref{eq:modEnt}) and (\ref{eq:aqEntropy}) reduce to statistics of dispersion about the mode in the binary case, but we omit the details for the sake of brevity.
\vspace{5mm}
\par\textbf{Index of qualitative variation}

The index of qualitative variation (IQV)~\cite{wilcox_indices_1973}, also known as the quadratic score~\cite{glasziou_test_1989}, is another common expression of uncertainty that is widely accepted in the social sciences~\cite{rubia_measures_2024}. It is defined in normalised form as
\begin{equation}
\label{eq:IQV}
    u_\mathrm{IQV}(\mathbf{p})=\dfrac{K}{K-1}\left(1-\sum_{k=1}^Kp_k^2\right),
\end{equation}
which for binary classification reduces to
\begin{equation}
    u_\mathrm{IQV}(\mathbf{p})=4\widehat{p}(1-\widehat{p}).
\end{equation}
The IQV is a special case of the $\alpha$-quadratic entropy in expression~\eqref{eq:aqEntropy} with $\alpha=1$.
\vspace{5mm}
\par\textbf{Coefficient of nominal variation}

A variant of the IQV proposed in~\cite{kvalseth_coefficients_1995} is the coefficient of nominal variation (CNV) defined as
\begin{equation}
\label{eq:CNV}
    u_\mathrm{CNV}(\mathbf{p})=1-\sqrt{1-u_\mathrm{IQV}(\mathbf{p})}=1-\sqrt{\dfrac{K\sum_{k=1}^Kp_k^2-1}{K-1}}.
\end{equation}
This transformation ensures that $u_\mathrm{CNV}(\mathbf{p})\leq u_\mathrm{IQV}(\mathbf{p})$ for all distributions, with equality for $\mathbf{p}^1$ and $\mathbf{p}_\text{unif}$. 
In terms of the normed distance $d(\mathbf{p},\mathbf{q})=||\mathbf{p}-\mathbf{q}||_2$ between two PMFs $\mathbf{p}$ and $\mathbf{q}$, the CNV can be rewritten as
\begin{equation}u_\mathrm{CNV}(\mathbf{p}) = 1 - \dfrac{d(\mathbf{p},\mathbf{p}_\text{unif})}{d(\mathbf{p}^1,\mathbf{p}_\text{unif})},\end{equation}
giving the CNV a natural interpretation in terms of the distance of the PMF from a uniform distribution.
For binary classification, expression~\eqref{eq:CNV} reduces to (see Appendix~\ref{app2}):
\begin{equation}
    \label{eq:CNVbin}
    u_\mathrm{CNV}(\mathbf{p})=2(1-\widehat{p}),
\end{equation}
which is equivalent to the WVR and the SDM.

\subsubsection{Comparing statistics of uncertainty}
\label{sec:compStat}
\begin{figure}[!ht]
    \centering
    \begin{subfigure}{0.5\linewidth}
        \includegraphics[width=\linewidth]{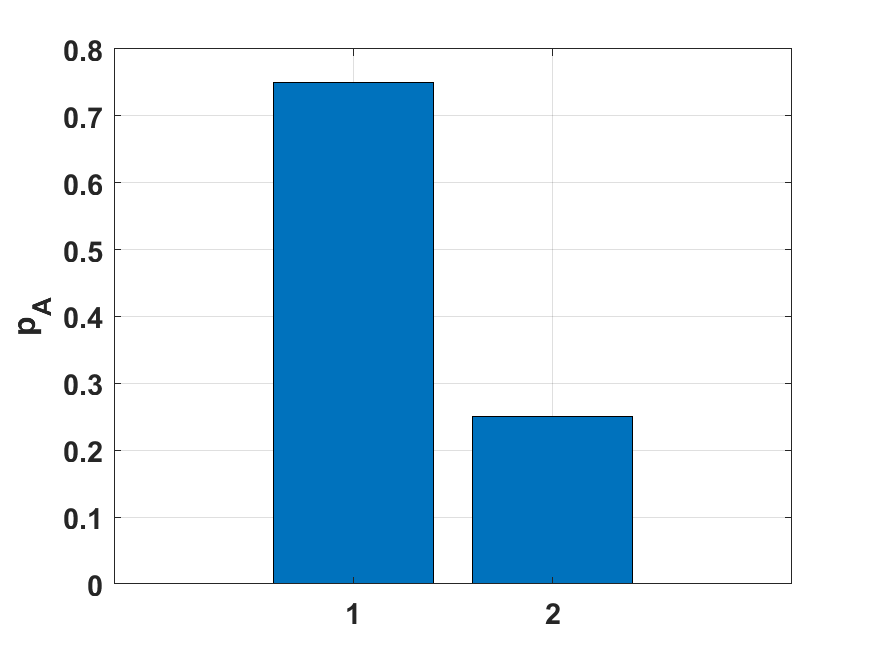}
        \caption{Example binary classification PMF: \mbox{$\mathbf{p}_A=[0.75,0.25]$}.}
    \end{subfigure}
    
    \begin{subfigure}{\linewidth}
        \includegraphics[width=0.45\linewidth]{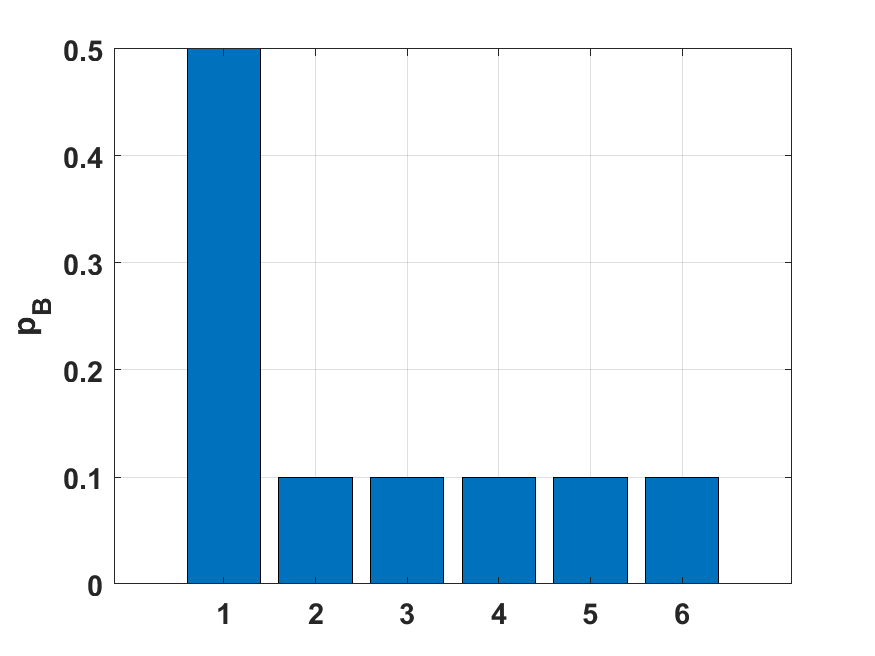}
        \includegraphics[width=0.45\linewidth]{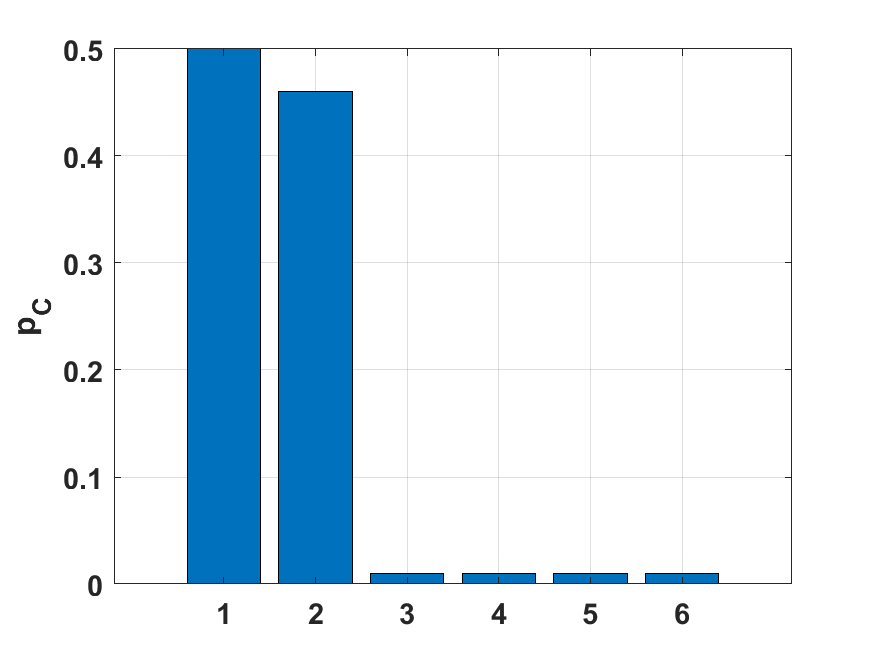}
        \caption{Example multi-class classification PMFs: \mbox{$\mathbf{p}_B=[0.5,0.1,0.1,0.1,0.1,0.1]$} and \mbox{$\mathbf{p}_C=[0.5,0.46,0.01,0.01,0.01,0.01]$}.}
    \end{subfigure}  
    \caption{Example predictive PMFs as output from ML classifiers.}
    \label{fig:comparePMFs}
\end{figure}

The statistics described in Section~\ref{sec:exUnc} have various advantages and disadvantages. In this section we compare their properties and performance on a few example predictive probability distributions. One particular example in binary classification is the PMF $\mathbf{p}_A$ shown in Figure~\ref{fig:comparePMFs}. Since $\mathbf{p}_A$ is halfway between total certainty and maximal uncertainty, i.e. $\widehat{p}_A=0.75$, it was suggested in~\cite{kvalseth_variation_2011} that the statistic of uncertainty should also be halfway between the extreme values, i.e., 0.5. We also consider the multi-class PMFs $\mathbf{p}_B$ and $\mathbf{p}_C$, which have the same model prediction $\widehat{y}$, with the same confidence (modal probability) of 0.5. The example PMFs are visualised in Figure~\ref{fig:comparePMFs} for comparison.

Table~\ref{tab:measComp} compares the various expressions of uncertainty for nominal properties. The first point to note is the result that $F(\mathbf{p}_B)\geq F(\mathbf{p}_C)$ for all $F$ considered, even though at first glance it might appear more obvious which is the best prediction of the classification model in $\mathbf{p}_B$. This result is due to the fact that the statistics are expressing the certainty that the mode can either be one of two possibilities in $\mathbf{p}_C$ (i.e. $(p_C)_1+(p_C)_2=0.96$), as the distribution is peaked about these two possible classes. Whereas for $\mathbf{p}_B$, the probability distribution is spread over all classes, leading to more uncertainty in the output class. 

If we expect the uncertainty of $\mathbf{p}_A$ to be 0.5, then we see that WVR, SDM and CNV satisfy this condition. In fact, WVR, SDM and CNV have the same expressions for binary classification (see Appendix~\ref{app2}). These three statistics also have the same values for $\mathbf{p}_B$ because expressions~\eqref{eq:SDM} and \eqref{eq:CNV} simplify to~\eqref{eq:WVR} when all the non-modal classes have the same probability (see Appendix~\ref{app2}). Thus, we only see a difference in values for $\mathbf{p}_C$, with WVR and CNV giving the highest and lowest values respectively.  

If we wish the statistic to use explicitly the modal probability (or confidence), as well as information about the entire distribution, then SDM is the only option. However, the SDM shows little difference in uncertainty between $\mathbf{p}_B$ and $\mathbf{p}_C$, where a larger difference from studying Figure~\ref{fig:comparePMFs} might be expected. In this case, CNV might be the more appropriate choice, but at the cost of no longer explicitly using information about the mode. 

\begin{table}[!ht]
    \caption{A comparison of expressions of uncertainty for nominal properties.}
    \label{tab:measComp}
    \centering
    \begin{tabular}{lcccccc}
        Statistic & Uses confidence & Uses full PMF & $\mathbf{p}_A$ & $\mathbf{p}_B$ & $\mathbf{p}_C$\\
        \hline
        
        WVR~\eqref{eq:WVR} & Y & N & 0.50 & 0.60 & 0.60\\
        
        UVR~\eqref{eq:UVR} & Y & N & 0.33 & 0.51 & 0.51\\
        
        SDM~\eqref{eq:SDM} & Y & Y & 0.50 & 0.60 & 0.56\\

        $H$~\eqref{eq:ShEnt} & N & Y & 0.81 & 0.84 & 0.50\\
        
        $H^*$~\eqref{eq:modEnt} & N & Y & 0.75 & 0.69 & 0.29\\
        
        IQV~\eqref{eq:IQV} & N & Y & 0.75 & 0.84 & 0.65\\
        
        CNV~\eqref{eq:CNV} & N & Y & 0.50 & 0.60 & 0.40\\
    \end{tabular}
    
\end{table}
\pagebreak
\subsection{Uncertainty propagation with nominal properties}\label{sec:pmfProp}

In Section~\ref{sec:exUnc}, we considered various ways in which the uncertainty in nominal properties can be expressed through a summary statistic. However, to propagate such uncertainty through a chain of measurement models, it is more appropriate to consider the full PMF. For quantitative properties, it is more difficult to propagate the full PDF analytically due to intractable integrals, and it is commonly the case that the full PDF is unknown and modelled using simpler forms such as assuming a Gaussian distribution. In such cases, summary statistics, such as the standard deviation, are propagated using the law of propagation of uncertainty~\cite{JCGMGUM}. Alternatively, where the PDF is well known, sampling techniques such as Monte Carlo (MC) can propagate the full PDF~\cite{JCGM101}.

In the case of a PDF having an analytical form, such as a Gaussian or rectangular distribution, the full PDF can be directly inferred from summary statistics, such as the standard deviation. On the other hand, the summary statistics for nominal properties described in Section~\ref{sec:exUnc} cannot be used to infer the PMF except in simplistic examples such as binary classification. It is therefore recommended to use the full PMF in uncertainty propagation through multistage measurement models that involve nominal properties.

Since, for nominal properties, the PMF can be defined explicitly, it can be used to obtain analytical results for propagation through measurement models. Consider an ML classification model $f$ with outputs $y\in\mathcal{C}_K$, which have a predictive distribution $\mathbf{p}=[p_1,\dots,p_K]$. Assume the outputs $y$ become inputs to the next measurement model $g$, which has both nominal and quantitative input variables, in the form of a $D$-dimensional vector $\mathbf{x}$, and outputs a quantitative variable $z\in\mathbb{R}$:
\begin{equation}\label{eq:genMdl}
    z=g(\mathbf{x},y),\quad \mathbf{x}\in\mathcal{X}\subset\mathbb{R}^D,\,y\in\mathcal{C}_K,\,z\in\mathbb{R}.
\end{equation}
Since there are $K$ possible values for $y$, we can describe $g$ as a model conditioned on a nominal variable, and therefore express $g$ in terms of $K$ models \mbox{$z = g(\mathbf{x},c_k) = g_k(\mathbf{x})$}, where $\mathbf{x}\in\mathcal{X}_k\subset\mathbb{R}^D$ for $k = 1, \dots, K$. This expression can be thought of as modelling a system which has $K$ different regimes, defined by the classes $c_k$.

The expected value of the output $z$ can be expressed in terms of the expected values of the functions $g_k(\mathbf{x})$:
\begin{equation}
    \mathbb{E}[z]=\sum_{k=1}^Kp_k\int_{\mathcal{X}_k}\mathrm{d}\mathbf{x}\,q_k(\mathbf{x})g(\mathbf{x},c_k)=\sum_{k=1}^Kp_k\int_{\mathcal{X}_k}\mathrm{d}\mathbf{x}\,q_k(\mathbf{x})g_k(\mathbf{x})=\sum_{k=1}^Kp_k\mu_k,
\end{equation}
where $q_k(\mathbf{x})$ are the PDFs of the input variable $\mathbf{x}\in\mathcal{X}_k$ and $\mu_k=\mathbb{E}_{\mathbf{x}\sim q_k(\mathbf{x})}[g_k(\mathbf{x})]$. The means $\mu_k$ can be evaluated in the standard way for quantitative input variables.

We can also compute the variance of the output $z$ in terms of summary statistics of the functions $g_k(\mathbf{x})$:
\begin{align}
    \text{Var}[z]&=\mathbb{E}[z^2]-\mathbb{E}[z]^2\nonumber\\
    &=\sum_{k=1}^Kp_k\mathbb{E}[g_k(\mathbf{x})^2]-\sum_{j,k=1}^Kp_jp_k\mu_j\mu_k\nonumber\\
    &=\sum_{k=1}^Kp_k(\sigma_k^2+\mu_k^2)-\sum_{j,k=1}^Kp_jp_k\mu_j\mu_k
\end{align}
where $\sigma_k^2=\text{Var}_{\mathbf{x}\sim q_k(\mathbf{x})}[g_k(\mathbf{x})]$ can be evaluated using the methods described in~\cite{JCGMGUM} for computing expressions of uncertainty in quantitative variables. 

For this simple example, the moments of the measurand given nominal property inputs can be determined analytically. More generally, MC techniques can be applied by sampling directly from the PMF to determine the output PDF $p(z)$. Proceeding in this way is relevant to measurement applications that depend upon a prior classification, such as in determining particle concentration from counting measurements~\cite{sipkens_tutorial_2023}.

\pagebreak
\section{Case studies}\label{sec:caseStudies}

In this section we present two applications of uncertainty evaluation using ML classification models: Land cover classification (Section~\ref{sec:landcover}), and atrial fibrillation detection using PPG signal data (Section~\ref{sec:AF}). In both cases we describe the background on the challenge, methods for obtaining class probabilities, and exploration of the contributions of different sources of uncertainty.

\subsection{Land cover classification}\label{sec:landcover}
\subsubsection{Background}
Land cover (LC) maps produced by classification of satellite imagery provide essential information on the Earth’s biophysical cover. This information serves as an important input into a variety of applications, e.g.
\begin{itemize}[noitemsep]
    \item Hydrological and climate modelling~\cite{Bohn2019,Wang2023}, where LC maps are used for setting up land surface conditions;
    \item National greenhouse gas accounting~\cite{IPCC2006}, where LC maps along with additional data are used for estimation of emissions associated with anthropogenic land use;
    \item Biodiversity monitoring~\cite{Skidmore2021}, where LC maps provide information on habitats and their fragmentation, etc.
\end{itemize}
Evaluation of uncertainties associated with LC maps is important both in its own right for the interpretation of such maps, and for the propagation of these uncertainties further -- to the downstream applications of LC maps mentioned above. As stated in Section~\ref{sec:pmfProp}, propagation of uncertainties of a nominal variable requires its full PMF. This PMF should include contributions of all uncertainty sources mentioned in Section~\ref{sec:uncEval} and illustrated in Figure~\ref{fig:uncertainty_sources}.

LC mapping is a multi-class classification problem, which currently is widely approached by ML. The most popular ML classification models used for LC mapping are tree-based models (such as random forests)~\cite{Kilcoyne2022,Marston2022,Friedl2022,Venter2021}, and convolutional neural networks (CNNs)~\cite{Brown2022,Karra2021}. There are many LC classification schemes. Some classification schemes include only a few, very broad, classes, e.g., tree cover, grassland, cropland, etc.~\cite{Zanaga2022}, whereas other classification schemes provide finer details, e.g. they distinguish broad-leaved and coniferous forest~\cite{EEA2020}. Some classification schemes have a nested structure where broad classes are further split into narrower classes~\cite{Friedl2022}. It is important to note that in the present case study we approach land cover classification in similar fashion to how it is commonly approached in the literature -- as a classification problem with a univariate output, i.e., each pixel considered separately. An important extension would be to perform classification with a multivariate output. In this case, classification would produce a collection of output variables (a collection of LC classes) corresponding to multiple pixels as opposed to each single pixel. Such an extension would make sense given how strongly correlated EO data typically is in the spatial domain.

\subsubsection{Data}
\label{sec:LCdata}
As previously mentioned in Section~\ref{subsec:ML}, a supervised ML classification model is trained on a data set consisting of pairs of several input variables and a categorical output. In our LC classification study, the input variables represented surface reflectance measurements performed in ten spectral bands of Sentinel-2 satellite over a single pixel. The output variable represented LC class in that pixel. Information on LC classes (i.e., LC labels) is typically obtained either manually (by visual interpretation of high/very high resolution satellite imagery or by ground surveys), automatically (e.g., based on pre-existing LC maps), or hybrid-like (following a combination of the two methods)~\cite{Moraes2024}. In our study, LC labels were obtained by an automatic collection procedure following a modified approach of UK Centre for Ecology and Hydrology (UKCEH), used for the collection of training data for the production of UK annual LC maps~\cite{Marston2022}. This method relies on the assumption that LC change is gradual. It uses UKCEH LC maps produced for three previous years to find pixels with stable LC classes (i.e. unchanged LC classes throughout these years). Further filtering was performed to identify such unchanged pixels where the RF classifier used by the UKCEH to produce their LC maps yielded a probability greater than 95 \%. In other words, we only retained pixels, where the LC label was assigned with the highest confidence. For the present case study, LC labels were collected over a $20\,\text{km}\times20\,\text{km}$ area of interest (AOI) in Scotland. The collection process yielded 189,142 unchanged pixels (i.e., 4.7 \% of the AOI) corresponding to forest, cropland, grassland and settlements classes.

The input variables (i.e., surface reflectance measurements) for the 189,142 unchanged pixels were retrieved from Sentinel-2 images collected over the AOI on the 1st of June, 2020 and 2021. Although space-borne surface reflectance measurements have many uncertainty sources associated with them, in the present study we choose to focus on one uncertainty source. This uncertainty source is related to the imperfect knowledge of aerosol optical depth (AOD) used in the atmospheric correction of the satellite imagery. It represents just one (although ordinarily dominant) contribution to the uncertainty sources (\ref{training data uncertainty}) and (\ref{prediction data uncertainty}) described in Section~\ref{sec:uncEval} and illustrated in Figure~\ref{fig:uncertainty_sources}. The AOD uncertainties were introduced into the atmospheric correction process by Monte Carlo (MC) sampling (we used 25 MC samples).The process yielded 25 realizations of surface reflectance measurements for each of the 189,142 unchanged pixels. The relatively low number of MC samples used in this study reflects its preliminary nature. The current implementation serves as a proof-of-concept, and we plan to increase both the number of MC samples and the range of uncertainty sources in future iterations of the framework.  See~\cite{bilson2025uncertaintyawarebayesianmachinelearning} for more details on the data used in this study.

\subsubsection{Propagation of uncertainties within Bayesian classification models}\label{sec:BayesUnc}
As illustrated in Figure~\ref{fig:uncertainty_sources}, ML classification models come with multiple sources of uncertainty. Most of these sources can be described in terms of probability distributions for particular parametric classification models (such as Na\"{i}ve Bayes, logistic regression, discriminant analysis~\cite{murphy2012machine} and neural networks~\cite{jospin_hands-bayesian_2022}) that naturally lend themselves to a Bayesian treatment. In this section we summarise a framework that can be followed to incorporate these sources of uncertainty for Bayesian generative ML classifiers, which explicitly model the input data uncertainties. For specific details of the framework and its implementation on land cover data, with reference to various uncertainty sources, see~\cite{bilson2025uncertaintyawarebayesianmachinelearning}.   

Let $\mathcal{D}=\{(\mathbf{x}_i,y_i)|i=1,\dots,N\}$ be the training data consisting of $N$ pairs of input and output observations, where $\mathbf{x}_i\in\mathcal{X}\subset\mathbb{R}^p$ and $y_i\in\mathcal{C}_K$. Then we can model the sources of uncertainty described in Figure~\ref{fig:uncertainty_sources} in the following way:  
\begin{itemize}
    \item The training data uncertainty (\ref{training data uncertainty}) from a measurement process can be described in terms of a probability distribution $p(\mathbf{x})$.
    \item The class assignment uncertainty of the ML classification task (\ref{ill-posed}) derives from the joint empirical probability distribution $p(\mathbf{x},y)$. This can be separated into two contributions via Bayes' rule
    \begin{equation}
        p(\mathbf{x},y=c_k)=p(\mathbf{x}|y=c_k)p(y=c_k),\quad k=1,\dots,K.
    \end{equation} 
    $p(\mathbf{x}|y=c_k)$ is the \emph{class conditional} probability distribution describing the distribution of the input data for each class. $p(y=c_k)$ is the probability distribution describing the prior class distribution (or \emph{class imbalance}) in the training data.
    \item ML models also come with model hyperparameters, which describe the choice of model (\ref{choice of model}), such as model architecture or optimisation method. Various hyperparameter optimisation tuning methods~\cite{franceschi2024hyperparameteroptimizationmachinelearning} can be used to determine the optimal model choice. Bayesian optimisation is one popular approach~\cite{NIPS2012_05311655}, and for such an approach, quantification of the uncertainties in choice of model has been attempted~\cite{pmlr-v151-tuo22a}.
    \item A parametric classification model can be described by a probability distribution $p(y_i|\mathbf{x}_i,\theta)$, where $\theta$ are the parameters of the model. The uncertainty associated with such a model is given by the probability distribution of the model parameters. The likelihood function $\mathcal{L}$ of the model is expressed in terms of the model distribution on the training data, and any additional training data uncertainty 
    \begin{equation}
        \mathcal{L}(\theta|\mathcal{D})=p(\mathcal{D}|\theta)=\prod_{i=1}^Np(\mathbf{x}_i,y_i|\theta)=\prod_{i=1}^Np(y_i|\mathbf{x}_i,\theta)p(\mathbf{x}_i).
    \end{equation}
    In a Bayesian setting, Bayes' rule is used to determine the posterior distribution in the model parameters (\ref{parameter uncertainty}) (otherwise known as \emph{Bayesian inference})
    \begin{equation}
        p(\theta|\mathcal{D})=\dfrac{p(\mathcal{D}|\theta)p(\theta)}{p(\mathcal{D})}.
    \end{equation}
    The Bayesian approach introduces an additional uncertainty $p(\theta)$, which is the prior distribution on the model parameters. This can be determined from expert knowledge, or one can use conjugate priors~\cite{murphy2012machine} if known for analytic determination of the posterior. $p(\mathcal{D})$ is known as the \emph{evidence}, and is typically intractable for more complex classification models such as CNNs. In such cases one must resort to sampling techniques such as Markov Chain Monte Carlo (MCMC)~\cite{brooks_handbook_2011}, approximation techniques such as variational inference~\cite{murphy2012machine} or MC dropout~\cite{kendall2017uncertainties} to estimate the posterior distribution. 
    \item Just as with training input data, new input data also comes with measurement uncertainty (\ref{prediction data uncertainty}). This can be described in terms of a probability distribution $p(\mathbf{x}_\mathrm{new})$.
    \item The probability distribution associated with the new input data uncertainty on an input $\mathbf{x}_\mathrm{new}$ (known as the \emph{posterior predictive}) can be evaluated as the expectation of the ML model probability of class $k$ with respect to the uncertainty in the model parameters $p(\theta|\mathcal{D})$:
    \begin{equation}        
    p(y_\mathrm{pred} = c_k|\mathbf{x}_\mathrm{new},\mathcal{D}) = \mathbb{E}_{\theta|\mathcal{D}}\left[P(y_{\mathrm{pred}} = c_k|\mathbf{x}_\mathrm{new},\theta)\right]
    \end{equation}
\end{itemize}
The posterior predictive distribution of the ML classification model can then be written in the form of a PMF with probability vector $\mathbf{p} = [p_1, \dots,p_K]^\top$, where $p_k := p(y_{\mathrm{pred}} = c_k|\mathbf{x}_{\mathrm{new}},\mathcal{D})$. This PMF can then by used to propagate uncertainties in further measurement models where the output of the ML classification model is used as an input, either analytically (as shown in Section~\ref{sec:pmfProp}), or through MC sampling of the PMF.
\subsubsection{Numerical examples}
\label{sec:LCnumEx}
We apply the Bayesian framework described in Section~\ref{sec:BayesUnc} to the datasets described in Section~\ref{sec:LCdata}, using the Bayesian quadratic discriminant analysis (BQDA) model presented in~\cite{bilson2025uncertaintyawarebayesianmachinelearning}. Specifically we compute the posterior predictive probabilities on test sets ($90\,\%$ of available pixels) from years 2020 and 2021, with training set taken from the remaining $10\,\%$ of pixels from year 2020. 

The confusion matrices (comparing class predictions and ``true" class labels) from taking the maximum probability as the predicted class are given in Figure~\ref{fig:LCcm}. We include the true/false positive and true/false negative rates to see how the model performs on each class, as well as computing the overall classification loss, which is the misclassification rate, or fraction of class predictions not from the true labelled class. 

For evaluation of the performance of the output class probabilities from the classification models, we use two expected values of proper scoring rules~\cite{gneiting2007strictly}, namely the multi-class cross-entropy loss (XE) and the Brier score (BS). Proper scoring rules are a principled way to assess the quality of the predictive probabilities produced by ML classification models~\cite{ferrer_analysis_2023}. 

Assume we have an ML classification model and a test dataset with $N$ observations (i.e. pixels in the case of land cover). For each observation $i=1,\dots,N$, the test dataset has a class label $k=1,\dots,K$. This can be represented by the binary matrix $y_{ik}$, where $y_{ik}=1$ if test observation $i$ has class label $k$, and zero otherwise. The output predictive probabilities of the model for each test observation $i$ and class label $k$ is represented by the probability matrix $p_{ik}$. The XE and BS for each test observation $i$ are then defined as
\begin{equation}
    \mathrm{XE}_i=-\sum_{k=1}^Ky_{ik}\log(p_{ik}), \qquad
    \mathrm{BS}_i=\frac{1}{K}\sum_{k=1}^K(y_{ik}-p_{ik})^2,
    \label{eq:PSR}
\end{equation}
The XE is similar to the information entropy defined in Section~\ref{sec:varStat} except that instead of only using the probability of being the most likely class from the model output, the output probabilities are used to calculate agreement to an independent test dataset, so that `predictive performance' is evaluated. The XE is commonly used in neural networks classifiers as the objective function to minimise during training. The BS is the mean squared difference between the test class label and the predicted probability. It considers all output probabilities, whereas XE only considers the probability associated with the test class label. 

To evaluate overall performance, we take expected values over the XE and BS defined in expression~\eqref{eq:PSR} for each test pixel $i$, as well as normalising the values such that 0 is perfect prediction, and 1 represents a predictive performance no better than using the prior class distribution of the test dataset as a trivial classifier (i.e. the fraction of land cover class abundance within the AOI). The expected XE (EXE) and BS (EBS) are defined as
\begin{equation}
    \mathrm{EXE}=\frac{1}{N}\sum_{i=1}^N\frac{\sum_{k=1}^Ky_{ik}\log(p_{ik})}{\sum_{k=1}^Kq_k\log(q_k)}, \qquad
    \mathrm{EBS}=\frac{1}{N}\sum_{i=1}^N\frac{\sum_{k=1}^K(y_{ik}-p_{ik})^2}{\sum_{k=1}^Kq_k(1-q_k)},
    \label{eq:EPSR}
\end{equation}
where $q_k$ is PMF of classes $k$ from the test dataset prior to ML classification.

\subsubsection{Results}
\begin{figure}[ht]
    \centering
    \begin{subfigure}{0.49\linewidth}
        \includegraphics[width=1\linewidth]{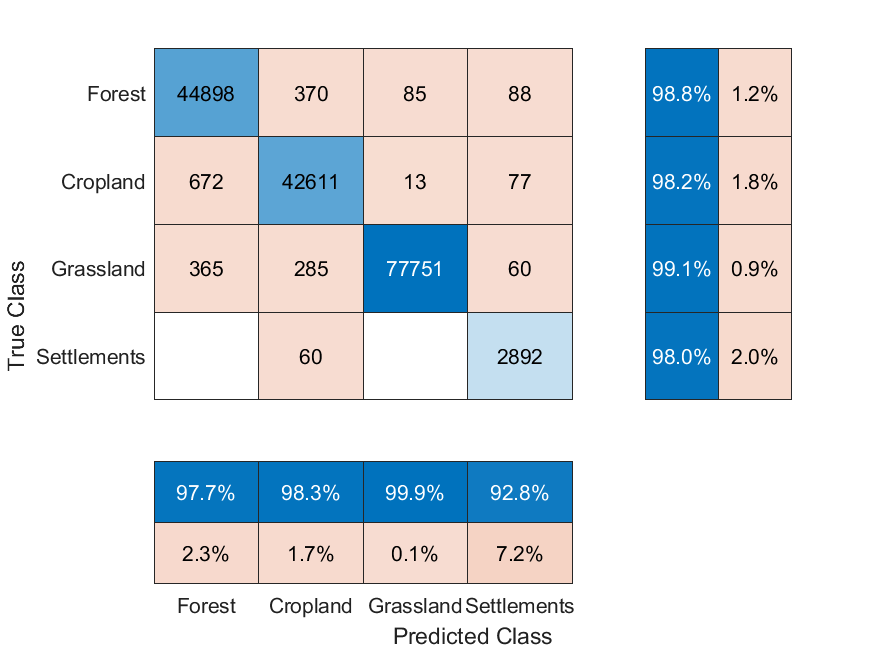}
        \caption{Test Year: 2020}
    \end{subfigure}   
    \begin{subfigure}{0.49\linewidth}
        \includegraphics[width=1\linewidth]{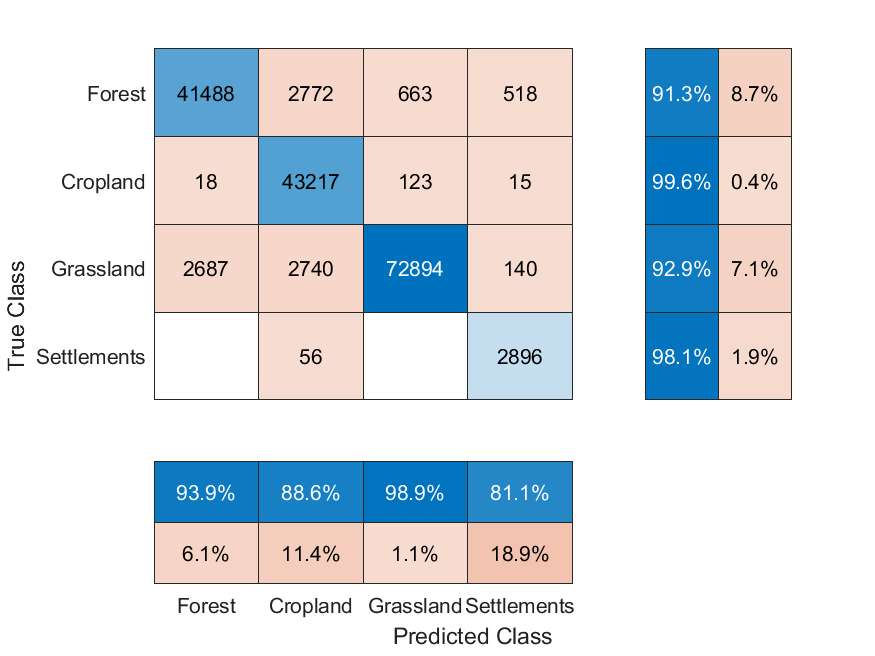}
        \caption{Test Year: 2021}
    \end{subfigure}
    \caption{Confusion matrices using BQDA. Final column corresponds to the true/false positive rates for each class. Final row corresponds to true/false negative rates.}
    \label{fig:LCcm}
\end{figure}

The resulting metric values for assessment of the performance of the classifiers are given in Table~\ref{tab:LCmetrics}. As is common with ML classification tasks, we see a degradation in performance when the model is tested on data outside the training domain (in this case a different year). We also observe from Figure~\ref{fig:LCcm} that the class `Settlements' have the highest false positive rate. This is in large part due to the low abundance of settlement pixels within the AOI, and so the ML classifier will tend to give other land cover classes with larger abundances a higher predictive probability.
\begin{table}[ht]
    \centering
    \begin{tabular}{c|ccc}
        Year & Loss & EXE & EBS \\
        \hline
        2020 & 0.012 & 0.079 & 0.031 \\
        2021 & 0.057 & 0.567 & 0.151 \\
    \end{tabular}
    \caption{Performance metrics for LC classification using BQDA model}
    \label{tab:LCmetrics}
\end{table}

In addition to the classification results, we also compute the expressions of uncertainty for nominal properties described in Section~\ref{sec:exUnc} using the posterior PMFs obtained within the Bayesian framework of~\cite{bilson2025uncertaintyawarebayesianmachinelearning}. Tables~\ref{tab:LCbw2020} and~\ref{tab:LCbw2021} show a summary of the behaviour in the chosen uncertainty statistics, for test years 2020 and 2021 respectively.
\begin{table}[ht]
    \centering
    \begin{tabular}{c|cccc}
        Statistic & Median/$10^{-8}$ & Mean/$10^{-3}$ & IQR/$10^{-5}$ & s.d./$10^{-2}$ \\
        \hline     
        WVR & $4.63$ & $8.10$ & $2.72$ & $5.00$\\
        UVR & $3.71$ & $6.48$ & $2.18$ & $4.00$\\
        SDM & $4.63$ & $7.81$ & $2.72$ & $4.72$\\
        $H$ & $88.8$ & $12.9$ & $24.6$ & $5.84$\\
        $H^*$ & $41.0$ & $7.35$ & $11.3$ & $3.63$\\
        IQV & $9.26$ & $12.4$ & $5.45$ & $6.73$\\
        CNV & $4.63$ & $7.04$ & $2.72$ & $4.02$\\
    \end{tabular}
    \caption{Summary of uncertainty statistics for LC classification. Test Year: 2020}
    \label{tab:LCbw2020}
\end{table}

\begin{table}[ht]
    \centering
    \begin{tabular}{c|cccc}
        Statistic & Median/$10^{-5}$ & Mean/$10^{-2}$ & IQR/$10^{-3}$ & s.d./$10^{-2}$\\
        \hline     
        WVR & $7.44$ & $3.03$ & $3.16$ & $9.94$\\
        UVR & $5.95$ & $2.42$ & $2.53$ & $7.95$\\
        SDM & $7.44$ & $2.91$ & $3.16$ & $9.34$\\
        $H$ & $51.9$ & $4.33$ & $13.2$ & $11.1$\\
        $H^*$ & $24.0$ & $2.56$ & $6.15$ & $7.17$\\
        IQV &$14.9$ & $4.48$ & $6.31$ & $13.1$\\
        CNV & $7.44$ & $2.59$ & $3.16$ & $7.97$\\
    \end{tabular}
    \caption{Summary of uncertainty statistics for LC classification. Test Year: 2021}
    \label{tab:LCbw2021}
\end{table}

In both tables, we see that the median and mean values are very small. This is due to the high accuracy or confidence of the LC classifiers, leading to low uncertainty. In both tables, we also see a pattern that the median in orders of magnitude smaller than the mean, and the IQR is orders of magnitude smaller than the s.d. This is because the distributions of the uncertainty statistics are highly skewed to the left, as again most of the classifier predictions were highly confident, leading values of the uncertainty statistics to be close to zero.

Note the large difference in scales between Tables~\ref{tab:LCbw2020} and~\ref{tab:LCbw2021}. This is due to the large classification performance degradation observed in the classification metrics from years 2020 to 2021 (see Figure~\ref{fig:LCcm}), causing the resulting predictive PMFs to be more uncertain. However, the uncertainty statistics still show high certainty in comparison to a uniform distribution. We also see a clear pattern for both years, with the information entropy $H$ having the largest median and IQR, and UVR having the smallest. Since the median and IQR represent the behaviour of the statistics for small values, the entropy is the most sensitive to variations in confident PMFs out of all the uncertainty statistics considered. UVR is the least sensitive in this case. Alternatively, the IQV has the largest s.d., and largest mean for year 2021, indicating that this statistic is more sensitive to variations in uncertain PMFs, i.e. PMFs that are close to uniform. The transformed entropy $H^*$ is the least sensitive, or most robust against small changes in uncertain PMFs. 

We observe that WVR, SDM, and CNV have equivalent median and IQR in Tables~\ref{tab:LCbw2020} and~\ref{tab:LCbw2021}. This is because the median and IQR are describing PMFs with a large $\widehat{p}$ in comparison with the other classes. This reduces these expressions for the uncertainty statistics of~\eqref{eq:SDM}\eqref{eq:CNV} to approximately~\eqref{eq:WVR}, essentially acting as a binary classification PMF where all other classes are treated as one. Appendix~\ref{app2} gives a derivation of this equivalence. 

\pagebreak
\subsection{Atrial fibrillation detection}\label{sec:AF}

\subsubsection{Background}

Atrial Fibrillation (AF) is a heart condition that causes an irregular and often fast heart rate. The global prevalence of AF was estimated to be 59 million individuals in 2019. Patients with AF are at increased risk of stroke, heart failure and dementia, and it is considered to be a 21$^\text{st}$ century cardiovascular disease epidemic~\cite{linz2024atrial}. 

Digital health solutions are often used for the timely detection of AF, and diagnosis of AF is typically based upon an ECG. Wearable or smartphone-based photoplethysmography (PPG) technology has more recently emerged as having clear potential to make continuous monitoring of AF more widely available. PPG measurements use a light source and a detector on the surface of the skin to measure changes in the optical properties of the skin’s microvascular bed caused by relative changes in blood volume~\cite{bench2024towards}.

Detection of AF from a time series segment of PPG measurements can be posed as a binary classification problem with two categories: \emph{positive} (AF is present) and \emph{negative} (AF is not present). Promising results for this classification task have already been reported. For example, a 122-patient study in~\cite{Wouters2024ComparativeEO} reported a specificity (Type I error) of 98.9 \% and a sensitivity (Type II error) of 100 \% for \emph{Fibricheck}'s PPG-based AF detection smartphone app~\cite{fibricheck}.

\subsubsection{Data}

We make use of a labelled PPG dataset called \emph{DeepBeat} introduced in~\cite{torres2020multi}. A custom subset of the dataset was selected to ensure a balanced label distribution and no overlap of patients between the training and test datasets. A total of 134 patients were selected, giving a derived dataset consisting of over 100,000 signal segments, with over 50,000 instances of both AF and non-AF. Each segment has a duration of 25 seconds and a sampling rate of 32 Hz. See~\cite{torres2020multi} for further details on the primary dataset, and see~\cite{bench2024towards} for further details on the derived dataset. 

\subsubsection{Probabilistic ML model}

There are generally two approaches to the AF classification task~\cite{charlton2022wearable}: (i) a rule-based classifier which uses features relating to pulse wave shape or inter-beat intervals; or (ii) an ML-based classifier which uses a training set of PPG data labelled by a clinical expert to optimise a generic empirical classification model (for example a neural network). This case study focuses on the latter approach.

Uncertainty evaluation is paramount in any classification model which informs medical diagnosis, where the consequences of misdiagnosis can be significant. Confidence in AF classification is affected by signal quality, and an evaluation of uncertainty is needed for clinicians to establish whether a classification can be used to reliably inform diagnosis. This case study therefore considers the use of a probabilistic ML classifier which, for each observation, outputs probabilities of being in each of the two classes. 

Recall that the dataset described in~\cite{torres2020multi} consists of signal segments from 134 patients. In~\cite{bench2024towards}, data from 110 of the patients was used to train the model, which allows for data from the remaining 24 patients to be used as examples of the use of the trained model for probabilistic AF classification. 

We give a brief overview of the ML model and its optimisation, and refer the interested reader to~\cite{bench2024towards} for further details. We use a CNN model which is a close variant of $\texttt{xresnet1d50}$ reported in~\cite{strodthoff2020deep}, which is itself a 1D variant of $\mathtt{xresnet}$~\cite{he2019bag}. A CNN model consists of a sequence of linear transformations (\emph{convolutions}) interspersed with simple non-linear operations, and it outputs prediction probabilities for each of the classes. Such a model is parametrised by a very large number of weights and biases defining each of the convolutions, and training the model involves optimising these parameters with respect to some loss function.

The probabilities returned by a CNN capture contributions to the prediction uncertainty arising from some of the sources of uncertainty mentioned in Section~\ref{sec:Introduction}, namely the class assignment uncertainty (\ref{ill-posed}), the training data uncertainty (\ref{training data uncertainty}) and the input test data uncertainty (\ref{prediction data uncertainty}). These sources of uncertainty are those which are properties of the classification task itself or the measurement system used to obtain the observations. In the ML community, such sources of uncertainty are often referred to as \emph{aleatoric}. Given observations, this is the contribution to the prediction uncertainty that cannot be reduced by choosing a better model or obtaining additional training samples. Aleatoric uncertainty is often distinguished in ML from \emph{epistemic} uncertainty, which is the uncertainty concerning the choice (\ref{choice of model}) and optimisation (\ref{parameter uncertainty}) of the model~\cite{hullermeier2021aleatoric}.

In order to capture in addition contributions to the prediction uncertainty due to model parameter uncertainty (\ref{parameter uncertainty}), an approach known as \emph{Monte Carlo Dropout}~\cite{gal2016dropout,kendall2017uncertainties} is used. The approach involves enforcing convolution weights to be zero at random with some fixed probability when both training and evaluating the model. A probability distribution is finally obtained which combines the aleatoric and epistemic contributions to the prediction uncertainty~\cite{kendall2017uncertainties}.

We note two sources of uncertainty which are inadequately captured by the Monte Carlo Dropout approach. Firstly, as is usually the case for uncertainty evaluation of ML models, uncertainty concerning the choice of model is ignored. It follows therefore that reliable uncertainty evaluation assumes the selection of an adequate type of model. Secondly, only uncertainty of the \emph{output} data is explicitly modelled in this approach, and uncertainty concerning \emph{input} data is not explicitly modelled. This is in contrast to the Bayesian QDA approach in Section~\ref{sec:landcover}. Approaches which explicitly model uncertainty of the inputs to deep learning models have been considered~\cite{martin2023aleatoric}, but the approach is computationally expensive and consequently not practically feasible for the large-scale AF detection task considered here. Full details of the involved steps and their linkage, and how aleatoric and epistemic uncertainty are captured, can be found in~\cite{bench2024towards}.

\subsubsection{Numerical examples}
\begin{figure}
    \centering
    \includegraphics[width=0.75\linewidth]{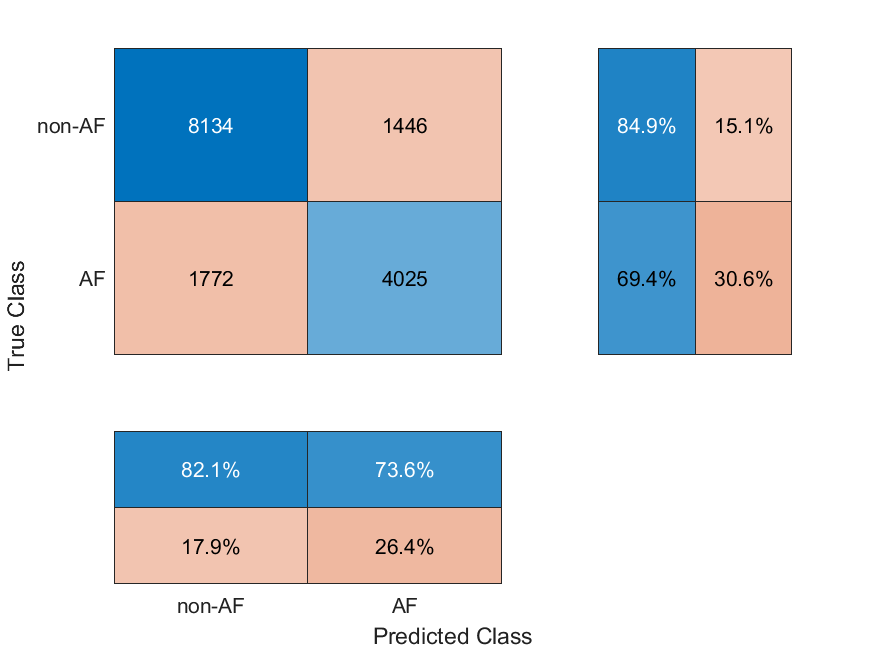}
    \caption{Confusion matrix for AF classifier on test dataset.}
    \label{fig:AFcm}
\end{figure}

We analyse the output predictive probabilities of the AF detection classifier using the same procedure as described in Section~\ref{sec:LCnumEx} for LC classification. The confusion matrix and summary statistics are given in Figure~\ref{fig:AFcm} and Table~\ref{tab:AFbw} respectively. 
The classification performance metrics are: classification loss = 0.209, EXE = 0.874, and EBS = 0.622.  Since the loss and other classification metrics are larger than in the case of LC classification, the uncertainty statistics in Table~\ref{tab:AFbw} are also larger than those in Tables~\ref{tab:LCbw2020} and~\ref{tab:LCbw2021}. 
\begin{table}
    \centering
    \begin{tabular}{c|cccc}
        Statistic & Median/$10^{-2}$ & Mean/$10^{-1}$ & IQR/$10^{-1}$ & s.d./$10^{-1}$\\
        \hline
        WVR & $6.98$ & $2.83$ & $5.87$ & $3.41$\\
        UVR & $4.66$ & $1.88$ & $3.92$ & $2.28$\\
        SDM & $6.98$ & $2.83$ & $5.87$ & $3.41$\\
        $H$ & $21.8$ & $4.01$ & $8.71$ & $4.10$\\
        $H^*$ & $16.3$ & $3.76$ & $8.30$ & $4.02$\\
        IQV & $13.5$ & $3.69$ & $8.29$ & $4.04$\\
        CNV & $6.98$ & $2.83$ & $5.87$ & $3.41$\\
    \end{tabular}
    \caption{Uncertainty statistics for AF classifier}
    \label{tab:AFbw}
\end{table}
The WVR, SDM and CNV give equivalent values as this is a binary classification problem, and their expressions are equivalent in this case (see Appendix~\ref{app2}). However, we observe a similar pattern to the uncertainty statistics computed for the land cover case study. Namely that the entropy is the most sensitive to variations in the predictive probability distributions, with UVR being the least sensitive.

\subsection{Summary of results}
The application of our ML classification uncertainty framework across different models and datasets allowed evaluation of the predictive PMFs, whilst taking into account the various sources of uncertainty described in Figure~\ref{fig:uncertainty_sources}. All uncertainty statistics of nominal properties described in Section~\ref{sec:varStat} were evaluated based on the outputted predictive PMFs, with results given in Tables~\ref{tab:LCbw2020},~\ref{tab:LCbw2021}, and Table~\ref{tab:AFbw}.

The results show consistent behaviour in the predictive PMFs, in that more accurate ML classification models produced uncertainty statistics with values close to zero. The results also showed that the information entropy was the most sensitive overall to variations in the PMF out of all uncertainty statistics considered, with UVR being the most robust. However, if one is more concerned with assessing highly uncertain PMFs, the IQV is the most sensitive. We also demonstrated that three uncertainty statistics (WVR, SDM, and CNV) are equivalent in the case of binary classification, and approximately equivalent for multi-class ML classification models with highly confident predictive PMFs.

\pagebreak
\section{Conclusion}\label{sec:conclusion}
In this paper, we have presented a metrological framework for evaluating uncertainty of nominal properties, motivated by the increasing use of ML classification models in a wide range of applications. We have identified and listed a number of sources of uncertainty within ML classification models which contribute to the output predictive uncertainty of the nominal property or variable, namely training data uncertainty, class assignment uncertainty, uncertainty about choice of model, model parameter uncertainty, and input data uncertainty, only some of which are currently addressed in the GUM suite of documents.

The work described in this paper has focused primarily on predictive uncertainty of ML classification models, with a specific focus on probabilistic ML classification models, i.e. models that return class probabilities for the output nominal variable. This focus included the evaluation of uncertainty for such nominal variables, which is not currently considered within the VIM and GUM. We have demonstrated that the PMF provides a complete characterisation of the uncertainty associated with nominal properties, and thus the predictive uncertainty in probabilistic ML classifiers. We have also motivated the use of the PMF of a nominal variable, as opposed to a summary statistic, in propagating such uncertainty through a multistage measurement model, representing either a metrological traceability chain or a processing chain for information products. We then provided example approaches using analytic or sampling methods of how to perform such propagation using the PMF. 

We have reviewed various statistics that have been proposed within the literature associated with expressing the predictive uncertainty of probabilistic ML classifiers. We provided a comparison between these statistics in terms of their fundamental properties, and applied them in the context of two case studies that rely upon a metrological analysis, namely land cover classification, and atrial fibrillation detection. We found that for both case studies, the information entropy, which is the most common statistic for uncertainty within the ML community, is in general, the most sensitive to variations in the predictive probability distributions. We have also shown that three of the uncertainty statistics (WVR, SDM and CNV) are equivalent for binary classification, and approximately equivalent in the multi-class case for confident ML classification model outputs. We have also illustrated how sources of uncertainty are addressed within the two case studies using different methodologies dependent on the classification modelling procedure.

The framework we have outlined builds upon and enhances previous work in uncertainty evaluation for nominal properties. The framework enables an extension of the GUM to uncertainty evaluation for nominal properties, and provides methods for handling the sources of uncertainty within ML classification, underpinning their use within metrological applications.

\section*{Acknowledgements}
The authors would like to thank Emma Woolliams (NPL) and Ian Smith (NPL) for providing helpful reviews and feedback. We would also like to thank Richard Brown, Xavier Loizeau, Samuel Hunt, and Bernardo Mota (all NPL) for their contributions and insights during the workshops held at NPL on 27$^\text{th}$ March and 27$^\text{th}$ November 2024.

\section*{Funding}
This work was funded by the UK Government's Department for Science, Innovation and Technology (DSIT) through the UK's National Measurement System (NMS) programmes. Anna Pustogvar’s work was also supported by a College of Science and Engineering (CSE) Scholarship from the University of Leicester.

\section*{Data Sources}
The original data used for LC classification study can be found at~\cite{Morton2020a,Morton2020b,Morton2020c,Morton2020d}. More details on how this data was processed for the LC classification study are provided in~\cite{bilson2025uncertaintyawarebayesianmachinelearning}. The original data for the case study on AF detection (\emph{DeepBeat}) can be found at~\cite{ashley2020deepbeat} and is described in~\cite{torres2020multi}. The custom data split used was described in~\cite{bench2024towards}.

\pagebreak
\printbibliography
\pagebreak
\appendix
\section{Derivation of WVR and UVR in terms of expected distance from mode}\label{app}
We can express the WVR and UVR defined in Section~\ref{sec:varStat} in terms of a simple distance metric from the mode(s), defined as
\begin{equation}
    d_k=\begin{cases}
        0 & p_k=\widehat{p},\\
        1 & \text{otherwise},
    \end{cases}\quad k=1,\dots,K.
\end{equation}
\vspace{5mm}
\par\textbf{Wilcox's Variation Ratio}

Using the fact that $\widehat{p}=1-\mathbf{d}\cdot\mathbf{p}$ for unimodal distributions, the expression for WVR~\eqref{eq:WVR} becomes
\begin{align}
    u_\text{WVR}(\mathbf{p})&=1-\frac{K\widehat{p}-1}{K-1}\nonumber\\
    &=1-\frac{K(1-\mathbf{d}\cdot\mathbf{p})-1}{K-1}\nonumber\\
    &=\frac{K}{K-1}\mathbf{d}\cdot\mathbf{p}
\end{align}
Since the probability vector $\mathbf{p}$ represents a PMF, we see that $\mathbf{d}\cdot\mathbf{p}=\mathbb{E}_\mathbf{p}[\mathbf{d}]$. This is the expected distance from the mode. Thus, we can express the normalised WVR in terms of the expected distance
\begin{equation}
    \label{eq:WVRdist}
    u_\text{WVR}(\mathbf{p})=\frac{K}{K-1}\mathbb{E}_\mathbf{p}[\mathbf{d}]
\end{equation}
We see a direct analogy with the variance for continuous variables in terms of the expected squared distance from the mean: $\text{Var}[X]=\mathbb{E}_p[d(X)^2]$, where $d(X)=X-\mathbb{E}_p[X]$.
\vspace{5mm}
\par\textbf{Universal Variation Ratio}

For multimodal distributions, $m\widehat{p}=1-\mathbb{E}_\mathbf{p}[\mathbf{d}]$, where $m$ is the number of modes. In this case, the UVR in expression~\eqref{eq:UVR} can be written as
\begin{align}
    u_\text{UVR}(\mathbf{p})&=\frac{K^2}{K^2-1}\left(1-\frac{\widehat{p}}{m}\right)\nonumber\\
    &=\frac{K^2}{K^2-1}\left(1-\frac{1-\mathbb{E}_\mathbf{p}[\mathbf{d}]}{m^2}\right)\nonumber\\
    &=\frac{K^2}{(K^2-1)m^2}\mathbb{E}_\mathbf{p}[\mathbf{d}]+\frac{K^2}{K^2-1}\frac{m^2-1}{m^2}
\end{align}
Thus the UVR can be expressed as a linear transformation of the expected distance from the mode even when the PMF is multi-modal, whereas expression~\eqref{eq:WVRdist} is only valid for unimodal distributions.

\section{Relationship between WVR, SDM and CNV}\label{app2}
Although expressions for the uncertainty statistics WVR~\eqref{eq:WVR}, SDM~\eqref{eq:SDM} and CNV~\eqref{eq:CNV} give different values in general, there is a relationship between them for certain forms of PMF $\mathbf{p}$.

Consider the unimodal case where the modal probability is $\widehat{p}$, and all other class probabilities are equal. This PMF can be written as
\begin{equation}\label{eq:exPMF}
    q_k=\begin{cases}
        \widehat{p} & c_k = \widehat{c},\\
        \frac{1-\widehat{p}}{K-1} & \text{otherwise},
    \end{cases}\quad k=1,\dots K.
\end{equation}

All binary class ($K=2$) PMFs can be written in this form. For multi-class classification models where outputs are confident in one class, the predictive output PMFs approximately follow the form of~\eqref{eq:exPMF}. 

We will now show that WVR, SDM and CNV uncertainty statistics are equivalent in such a scenario. Substituting~\eqref{eq:exPMF} into~\eqref{eq:SDM} leads to
\begin{align*}
    u_\text{SDM}(\mathbf{q})&=1-\sqrt{\dfrac{(K-1)(\widehat{p}-\frac{1-\widehat{p}}{K-1})^2}{K-1}}\\
    &=1-\widehat{p}+\frac{1-\widehat{p}}{K-1}\\
    &=1-\frac{K\widehat{p}-1}{K-1}=u_\text{WVR}(\mathbf{q}).
\end{align*}
Now substituting~\eqref{eq:exPMF} into~\eqref{eq:CNV} yields
\begin{align*}
    u_\text{CNV}(\mathbf{q})&=1-\sqrt{\frac{K\left(\widehat{p}^2+(K-1)\left(\frac{1-\widehat{p}}{K-1}\right)^2\right)-1}{K-1}}\\
    &=1-\frac{1}{K-1}\sqrt{(K-1)(K\widehat{p}^2-1)+K(1-\widehat{p})^2}\\
    &=1-\frac{1}{K-1}\sqrt{K^2\widehat{p}^2-2K\widehat{p}+1}\\
    &=1-\frac{1}{K-1}\sqrt{(K\widehat{p}-1)^2}=u_\text{WVR}(\mathbf{q})
\end{align*}
This also explains the equivalence in the binary class uncertainty expressions~\eqref{eq:WVRbin},~\eqref{eq:SDMbin}, and~\eqref{eq:CNVbin}.
\end{document}